\documentclass{article}

\PassOptionsToPackage{numbers,compress}{natbib}
\usepackage[preprint]{neurips_2025}
\usepackage[utf8]{inputenc}
\usepackage[T1]{fontenc}

\usepackage{amsmath, amssymb, amsfonts, amsthm}
\usepackage{graphicx}
\usepackage{wrapfig}
\usepackage{lipsum} 
\usepackage{subcaption}
\usepackage{booktabs}
\usepackage{booktabs} 
\usepackage{amsmath}  
\usepackage{array}    
\usepackage{tabularx}
\usepackage{etoc} 
\addtocontents{toc}{\protect\setcounter{tocdepth}{-1}}

\usepackage{array}
\usepackage{multirow}
\usepackage{caption}

\usepackage{algorithm}
\usepackage{algpseudocode}

\usepackage{microtype}
\usepackage{nicefrac}
\usepackage{enumitem}
\usepackage{comment}
\usepackage{tcolorbox}
\usepackage{xcolor}
\usepackage{hyperref}
\usepackage{url}

\definecolor{morange}{HTML}{FF9800}
\definecolor{daleviolet}{HTML}{9C27B0}
\definecolor{lightblue}{RGB}{220,235,255}
\definecolor{lightgreen}{RGB}{220,255,220}
\definecolor{lightorange}{RGB}{255,240,220}
\definecolor{lightpurple}{RGB}{240,230,255}
\definecolor{lightgray}{gray}{0.9}
\definecolor{burntorange}{RGB}{230,159,0}
\definecolor{softorange}{RGB}{251,193,94}
\definecolor{cobaltblue}{RGB}{0,114,178}
\definecolor{charcoal}{RGB}{0,0,0}
\definecolor{darkgreen}{RGB}{0,158,115}
\definecolor{mintgreen}{RGB}{102,221,170}
\definecolor{magenta}{RGB}{204,121,167}
\definecolor{peach}{RGB}{244,165,193}
\definecolor{brickred}{RGB}{213,94,0}
\definecolor{slategray}{RGB}{153,153,153}
\definecolor{threedblue}{RGB}{0,0,139}
\definecolor{threedgreen}{RGB}{0,100,0}
\definecolor{threedgray}{RGB}{128,128,128}
\definecolor{pdblue}{HTML}{2196F3}
\definecolor{alegreen}{HTML}{4CAF50}
\definecolor{a2d2ered}{HTML}{F44336}

\newtheorem{theorem}{Theorem}

\newtheorem{corollary}{Corollary}

\theoremstyle{definition}
\newtheorem{definition}{Definition}
\newtheorem{assumption}{Assumption}

\title{Accumulated Aggregated D-Optimal Designs for Estimating Main Effects in Black-Box Models}

\author{
  Chih-Yu Chang \\
  Department of Mathematics \\
  Imperial College London \\
  London, United Kingdom \\
  \texttt{c.chang25@imperial.ac.uk}
  \And
  Ming-Chung Chang \\
  Institute of Statistical Science \\
  Academia Sinica \\
  Taipei, Taiwan \\
  \texttt{mcchang0131@as.edu.tw}
}

\begin{document}
\maketitle
\begin{abstract}
Estimating how individual input variables affect the output of a black-box model is a central task in explainable machine learning. However, existing methods suffer from two key limitations: sensitivity to out-of-distribution (OOD) evaluations, which arises when query points are placed far from the data manifold, and instability under feature correlation, which can lead to unreliable effect estimates in practice. We introduce a unified view of main effect estimation as a $\emph{design problem}$, which reveals that all existing methods differ only in their choice of evaluation locations. Building on this formulation, we propose A2D2E, an \textbf{E}stimator based on \textbf{A}ccumulated \textbf{A}ggregated \textbf{D}-Optimal \textbf{D}esigns, which replaces evaluations with a $D$-optimal hypercube design to minimize the variance of main effect estimation. A2D2E is model-agnostic, requires no differentiability of the predictor, and admits a closed-form estimator with complexity comparable to existing approaches. We establish that A2D2E is consistent to the same population target as ALE, and extend this result to the realistic setting where only a surrogate model is available. Through extensive simulations across multiple predictive models and dependence settings, we demonstrate that A2D2E outperforms ALE-based methods, with the largest gains under high feature correlation.
\end{abstract}

\begin{figure}[h]
    \centering
    \includegraphics[width=0.9\textwidth]{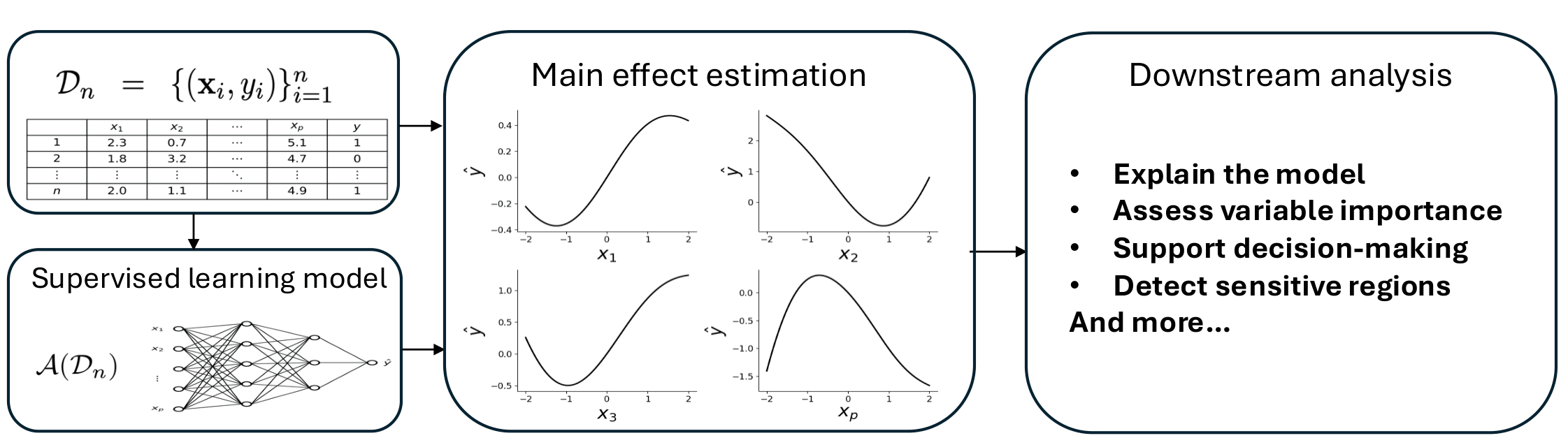}
    \caption{\textbf{Procedure and practical aspect of main effect estimation.} Given observed data and a fitted supervised learning model, the goal of main effect estimation is to recover interpretable effect curves that describe how individual inputs affect the predicted response.}
    \label{fig:intro_main_effect}
\end{figure}
 
\section{Introduction}
With the increasing availability of large datasets and computing resources, complex predictive models have become a major research focus. However, these models often sacrifice interpretability compared to traditional models such as linear or logistic regression. As illustrated in Fig.~\ref{fig:intro_main_effect}, a growing number of real-world applications require a more specific goal: \textbf{estimating the effect of individual variables on the predicted response}, which enables downstream tasks such as explaining the model, assessing variable importance, supporting decision-making, and detecting sensitive regions. Indeed, main effect estimation has been widely adopted across diverse domains, including environmental and natural sciences \citep{welchowski2022techniques, gholami2024assessment, tella2026advancing}, engineering and physical sciences \citep{roy2025prediction, fathipour2023mean}, and machine learning \citep{brenning2023interpreting, moosbauer2021explaining}.

Despite its practical value, existing effect estimation methods 
have well-documented limitations. PD plots produce misleading 
results under feature correlation~\cite{shi2023clarifying}, 
and the M plot suffers from similar issues~\cite{apley2020visualizing}. 
While ALE improves robustness to correlated variables, it may 
suffer from scalability issues in high-dimensional settings and 
relies on within-bin sampling that can lead to inaccuracies under 
OOD scenarios \cite{gkolemis2023dale}. DALE~\cite{gkolemis2023dale} 
addresses the latter by avoiding unseen data points, but requires 
the predictor to be differentiable, excluding nonparametric models 
such as $\mathcal{K}$-nearest neighbors ($\mathcal{K}$-NN), as well as black-box predictors. 
This motivates the need for a general approach that is both stable 
regardless of feature correlation and applicable to any predictive 
model, including non-differentiable ones.

\paragraph{Main contributions.} We address this need with the following contributions. First, we introduce a novel view of main effect estimation as a \emph{design problem}, providing a unified formulation that encompasses existing methods such as PD and ALE as special cases (Sec.~\ref{sec:design}). Building on this formulation, we propose A2D2E in Sec.~\ref{sec:a2d2e}. The proposed method integrates concepts from experimental design, which guarantees minimum variance estimation of the local slope under a correctly specified local linear model within each neighbourhood. Theoretically, we prove the consistency property of A2D2E, supporting its reliability in the realistic setting where only a surrogate $\hat{f}$ is available. Finally, we conduct extensive simulations across multiple prediction models and dependence settings in Sec.~\ref{sec:exp}. A sensitivity analysis on the required hyperparameters 
is conducted to verify the stability of the proposed method.
\subsection{Related to Previous Works}

Our unified formulation reveals that main effect estimation reduces at its core to local slope estimation, connecting it to a rich statistical literature on nonparametric derivative estimation \citep{muller1987bandwidth, gasser1984estimating, 
zhou2000derivative, charnigo2011generalized}, finite difference methods \citep{de2013derivative, knowles2014methods, ahnert2007numerical}, and optimal design for slope estimation \citep{muller1984optimal, 
chernoff1953locally}. As these works have developed largely independently of the interpretability literature, we defer their review to Appendix~\ref{app:relatedwork} and focus here on direct competitors in main effect estimation. Beyond PD, M, and ALE, related approaches include ICE plots \cite{goldstein2015peeking}, functional ANOVA \cite{hooker2007generalized}, and SHAP \cite{lundberg2017unified}, which target individual-level effects, variance decomposition, and scalar importance scores respectively. We focus on PD, M, ALE, and DALE as direct competitors, since they share the same estimation target, a continuous main effect curve, differing only in marginalization and evaluation location choice.

\section{Proposed Method}\label{sec2}

\paragraph{Notations.} Let $\hat{f} = \mathcal{A}(\mathcal{D}_n)$ 
denote a prediction model trained on a dataset 
$\mathcal{D}_n = \{(\mathbf{x}_i, y_i)\}_{i=1}^n$ 
using a supervised learning algorithm $\mathcal{A}$, 
where $\mathbf{x}_i \in \mathbb{R}^p$ and $y_i \in 
\mathbb{R}$. We use upper case to denote the random variable. For $d \in [p]:=\{1,...,p\}$, we denote the $d$-th 
coordinate of $\mathbf{x}_i$ as $x_{i,d}$ and the 
remaining coordinates as $\mathbf{x}_{i,\backslash d} 
\in \mathbb{R}^{p-1}$. The goal is to estimate the 
main effect function $f_d : \mathbb{R} \to \mathbb{R}$ 
for each $d \in [p]$, which describes how $\hat{f}$ 
varies with $x_d$ marginally. We focus on main 
effect functions as they are the primary tool for applications 
\citep{zhu2025unveiling, hakkoum2024global}, 
while higher-order effects remain limited in both 
methodology and evaluation \citep{apley2020visualizing, 
gkolemis2023dale}.  

Our central observation is that estimating $f_d$ 
is fundamentally a \emph{design problem}: existing 
methods (PD and ALE) select locations at 
which to evaluate $\hat{f}$, yet none does so with 
an explicit optimality criterion. We formalize this 
view in Sec.~\ref{sec:design}, which serves as 
the foundation for the proposed A2D2E, introduced 
in Sec.~\ref{sec:a2d2e}. Theoretical properties of the proposed method are presented in Sec.~\ref{sec:thm}.

\subsection{A Unified View: Effect Estimation as a Design Problem}\label{sec:design}

We claim that estimating $f_d$ of a black-box 
model can be formulated as a \emph{design problem}. In classical experimental design, one seeks to estimate a target quantity by selecting a set of input locations (the \emph{design}) at which to query a response function, with the goal of maximizing the statistical efficiency of the resulting estimator \citep{kiefer1959optimum, wu2011experiments}. The main effect estimation problem fits naturally into this framework, as we elaborate below.

Common approaches for estimating the main effect of dimension $d$ can be written as an accumulation of local increments over a partition $\{[z_d^k, z_d^{k+1}]\}_{k=1}^K$ of the support of $X_d$:
\begin{equation}\label{eq:accumulation}
\hat{f}_d(t) = \sum_{k=1}^{J(t)} \hat{\Delta}_d^k + c,
\end{equation}
where $c$ is an arbitrary constant that is 
typically chosen so that $\hat{f}_d$ has mean 
zero over the evaluated location (e.g., 
$c=\frac{1}{K}\sum_{k=1}^K \hat{f}_d(z_d^{k+1})$), 
$J(t)$ is the index of the bin containing $t$, 
$\hat{\Delta}_d^k$ is the estimated increment in 
bin $k$, and $K$ is the number of bins. Estimating 
$\hat{f}_d$ therefore reduces to estimating each 
$\hat{\Delta}_d^k$ separately. This is analogous 
to piecewise local regression or regression splines 
\citep{hastie2017generalized}, where a global 
function is reconstructed by stitching together 
locally fitted pieces. Here, each $\hat{\Delta}_d^k$ 
plays the role of a local contribution, and the 
accumulation recovers the full effect curve.

For each bin $k$, define the index set $I_d^k\subseteq\{1,...,n\}$ which is the data that are used to estimate $\Delta_d^k$. Our key observation is that the estimation of the bin increment $\Delta_d^k$ can be decomposed into two levels. First, for each observation $i \in I_d^k$, we estimate a \emph{local slope} in 
the $d$-th direction by fitting a local linear approximation of $\hat{f}$ around $\mathbf{x}_i$. Second, we aggregate these local slopes across all 
observations in the bin. Thus, the bin-level slope is an average over the conditional distribution of $X_{\backslash d}$ given $X_d \in [z_d^k, z_d^{k+1}]$.

Specifically, for each $i \in I_d^k$, we consider the local linear approximation
\begin{equation}\label{eq:local_linear}
\hat{f}(\mathbf{u})
\approx
\alpha_{k,i}
+
\beta_{k,i}(u_d-x_{i,d})
+
\boldsymbol{\gamma}_{k,i}^{\top}
(\mathbf{u}_{\backslash d}-\mathbf{x}_{i,\backslash d}),
\qquad \mathbf{u} \in \mathcal{S}_i,
\end{equation}
where $\mathcal{S}_i \subset \mathbb{R}^p$ is a local design set around 
$\mathbf{x}_i$ and $\beta_{k,i}$ is the local slope in the $d$-th coordinate. 
When two evaluation points share the same $\mathbf{u}_{\backslash d}$, 
the intercept and nuisance terms cancel, so the bin-level slope 
reduces to an average of local slopes:
\begin{equation}\label{eq:aggregation}
\beta_{k}
=
\frac{1}{|I_d^k|}
\sum_{i\in I_d^k} \beta_{k,i},
\end{equation}
and the bin increment follows as $\Delta_d^k=(z_d^{k+1}-z_d^k)\,\beta_{k}$.
Estimating $\Delta_d^k$ therefore decomposes into 
\textbf{(a)} choosing the local design $\mathcal{S}_i$ 
to estimate each $\beta_{k,i}$, and 
\textbf{(b)} choosing the index set $I_d^k$ to determine 
how local slopes are aggregated, which varies according 
to the estimand targeted by each method.

PD plot sets $I_d^k = \{1,\dots,n\}$ and 
constructs $
\mathcal{S}_i
=
\{(z_d^k,\mathbf{x}_{i,\backslash d}),
(z_d^{k+1},\mathbf{x}_{i,\backslash d})\}.$ Using the two-point local design gives
\[
\hat{\beta}_{k,i}
=
\frac{
\hat{f}(z_d^{k+1},\mathbf{x}_{i,\backslash d})
-
\hat{f}(z_d^k,\mathbf{x}_{i,\backslash d})
}{
z_d^{k+1}-z_d^k
},
\qquad
\hat{\beta}_{k}
=
\frac{1}{n}\sum_{i=1}^n \hat{\beta}_{k,i}.
\]
Since $I^k_d$ includes all observations, $\beta_k$ marginalizes 
over the full distribution of $X_{\setminus d}$. However, when $X_d$ is 
correlated with $X_{\setminus d}$, the artificial evaluation points in 
$S_i$ may lie far from the data manifold, making $\hat{\beta}_{d,k}$ 
susceptible to extrapolation.

The ALE plot improves PD plot by restricting the 
aggregation to observations inside the bin, namely 
$I_d^k=\{i:x_{i,d}\in[z_d^k,z_d^{k+1}]\}$, and constructing $\mathcal{S}_i
=
\{(z_d^k,\mathbf{x}_{i,\backslash d}),
(z_d^{k+1},\mathbf{x}_{i,\backslash d})\},
\ i\in I_d^k.$
This gives
\[
\hat{\beta}_{k,i}
=
\frac{
\hat{f}(z_d^{k+1},\mathbf{x}_{i,\backslash d})
-
\hat{f}(z_d^k,\mathbf{x}_{i,\backslash d})
}{
z_d^{k+1}-z_d^k
},
\qquad
\hat{\beta}_{k}=\frac{1}{|I_d^k|}
\sum_{i\in I_d^k}\hat{\beta}_{k,i}.
\]

It is worth noting that although ALE also constructs two artificial points 
for each observation, the maximum distance (measured in $\ell_2$ norm) 
between any point in $\mathcal{S}_{\mathrm{ALE},i}$ and the observed data 
$\mathcal{D}_n$ is upper bounded by $z_d^{k+1}-z_d^k$, since $\mathbf{x}_i$ 
itself lies in the $k$-th bin. Unlike PD, ALE restricts aggregation to 
in-bin observations, reducing extrapolation risk, but retains the same 
two-point design. With $|\mathcal{S}_i| = 2$, the system is exactly 
determined with no redundancy for variance reduction, and no existing 
method selects $\mathcal{S}_i$ using an explicit optimality criterion. 
A richer local design $\mathcal{S}_i$ could therefore yield a more 
efficient estimator of the local slope $\beta_{k,i}$, and hence a 
lower-variance estimator of the aggregated bin-level slope $\beta_k$. 
A2D2E exploits this by replacing the heuristic two-point design 
in each local regression with a $D$-optimal design, while keeping 
the aggregation step~\eqref{eq:aggregation} and the index set 
$I^k_d$ identical to ALE.

\subsection{A2D2E}
\label{sec:a2d2e}

As established in 
Sec.~\ref{sec:design}, existing methods rely on 
exactly two evaluation points per observation, 
admitting no redundancy for variance reduction. 
A2D2E addresses this by constructing a local 
design $\mathcal{S}_i$ around each $\mathbf{x}_i$ 
according to $D$-optimal design theory while keeping $I_d^k=\{i:x_{i,d}\in[z_d^k,z_d^{k+1}]\}$ the same as ALE.
\paragraph{D-optimal local regression.}
Given the local linear model~\eqref{eq:local_linear}, 
estimating $\beta_{k,i}$ at observation $\mathbf{x}_i$ 
requires evaluating $\hat{f}$ at a set of input locations 
$\mathcal{S}_i$ and fitting an OLS regression. A natural 
question is how to choose $\mathcal{S}_i$: one could simply 
sample a large number of random locations around $\mathbf{x}_i$, 
but this is statistically inefficient and computationally 
wasteful, as many of the sampled points may carry redundant 
information about $\beta_{k,i}$. Instead, we seek a design, denoted by $\mathcal{S}_{\text{A2D2E},i}$, of fixed size that maximizes the statistical efficiency 
of the OLS estimator of $\beta_{k,i}$.

Evaluating $\hat{f}$ at all points in $\mathcal{S}_{\text{A2D2E},i}$ and 
stacking the responses gives a linear system $\mathbf{y}_i = V_i \boldsymbol{b}_i 
+ \boldsymbol{\varepsilon}_i$, where $\mathbf{y}_i $ is the 
vector of model evaluations at the design points, $V_i$ is the design matrix 
whose rows correspond to the points in $\mathcal{S}_{\text{A2D2E},i}$, 
$\boldsymbol{b}_i := (\alpha_{k,i}, \beta_{k,i}, \boldsymbol{\gamma}_{k,i})^\top$, and $\boldsymbol{\varepsilon}_i \in \mathbb{R}^{2^p}$ is the vector of local linear approximation errors. For the purpose of deriving the $D$-optimal design, we treat these as i.i.d.\ with mean zero and variance $\sigma^2$. In practice they reflect the discrepancy between 
$\hat{f}$ and its local linear approximation, which is of order $O(\delta^2)$ by a second-order Taylor expansion (Details can be found in Appendix~\ref{app:proof_prop}). Under this assumption, the OLS estimator $\hat{\boldsymbol{b}}_i = (V_i^\top V_i)^{-1} V_i^\top \mathbf{y}_i$ has covariance matrix $\sigma^2(V_i^\top V_i)^{-1}$, so minimizing estimation variance amounts to making $(V_i^\top V_i)^{-1}$ small in a matrix sense. $D$-optimality \citep{kiefer1959optimum} formalizes this by maximizing 
$\det(V_i^\top V_i)$, which minimizes the volume of the confidence ellipsoid for $\hat{\boldsymbol{b}}_i$ and ensures every evaluation of $\hat{f}$ contributes 
maximally to estimating $\boldsymbol{b}_i$. For the local linear model with $p$ input variables and design points constrained to a neighborhood of radius $\delta/2$ around $\mathbf{x}_i$, the $D$-optimal design is given by the 
$2^p$ vertices of the hypercube centered at $\mathbf{x}_i$ with edge length $\delta$ \citep{wu2011experiments},
\begin{equation}
\label{eq:si_selection}
\mathcal{S}_{\text{A2D2E},i} = \left\{ \mathbf{x}_i + 
\frac{\delta}{2}s \;\middle|\; 
s \in \{-1,+1\}^p \right\}.
\end{equation}

It is worth noting that although A2D2E constructs $2^p$ artificial evaluation 
points per observation, the maximum $\ell_2$ distance between any hypercube vertex and 
its anchor point $\mathbf{x}_i$ is $\frac{\delta\sqrt{p}}{2}$. A2D2E therefore has a 
smaller maximum displacement than ALE whenever $\delta < \frac{2(z_d^{k+1}-z_d^k)}{\sqrt{p}}$, 
reducing sensitivity to OOD evaluations in this regime. We further verify that the 
estimator is not sensitive to the choice of small $\delta$ in Sec.~\ref{sec:exp}.

To verify optimality, note that the local linear 
model~\eqref{eq:local_linear} is parameterized in terms 
of deviations $(\mathbf{u} - \mathbf{x}_i)$, so once 
$\mathcal{S}_{\text{A2D2E},i}$ is substituted, the rows 
of $V_i$ take the centered form $\left(1, 
\frac{\delta}{2}[s]_d, \frac{\delta}{2} 
s_{\backslash d}^\top\right)$ for $s \in \{-1,+1\}^p$, 
where the first column corresponds to $\alpha_{k,i}$, 
the second to $\beta_{k,i}$, and the remaining columns 
to $\boldsymbol{\gamma}_{k,i}$. Since the hypercube 
vertices are symmetric around $\mathbf{x}_i$, the 
intercept column is orthogonal to all other columns, 
and the Gram matrix of the non-intercept block 
$\tilde{V}_i$ satisfies 
$\tilde{V}_i^\top \tilde{V}_i = 2^{p-2}\delta^2 I_p$, 
since the columns of $\tilde{V}_i$ are mutually 
orthogonal with squared norm $2^{p-2}\delta^2$. This 
gives $\det(\tilde{V}_i^\top \tilde{V}_i) = 
(2^{p-2}\delta^2)^p$, the maximum achievable under the 
constraint that each coordinate lies within 
$[-\delta/2, \delta/2]$ of $\mathbf{x}_i$ \cite{kiefer1959optimum,wu2011experiments}. 


The slope estimate can also be simplified to the 
scalar operation
\begin{equation}
\label{eq:estimate_beta}
\hat{\beta}_{k,i} = [\hat{\boldsymbol{b}}_{i}]_d =  
\frac{1}{2^{p-2}\delta^2}
\sum_{s \in \{-1,+1\}^p} 
\frac{\delta}{2} [s]_d \cdot 
\hat{f}\!\left(\mathbf{x}_i + 
\frac{\delta}{2}s\right),
\end{equation}


We note that these optimality results rely on the local linear model~\eqref{eq:local_linear} being correctly specified within the neighborhood of radius $\delta/2$, which holds approximately when $\delta$ is small relative to the length scale of $\hat{f}$. Under misspecification, the estimator remains consistent as $\delta \to 0$ by Theorem~\ref{thm:thm1}, though variance optimality no longer holds exactly.


\paragraph{Aggregation and accumulation.}
The per-observation slope estimates are aggregated 
within each bin and scaled by the bin width to 
obtain the increment estimate $\hat{\Delta}_d^k = (z_d^{k+1} - z_d^k) \cdot 
\frac{1}{|I_d^k|} \sum_{i \in I_d^k} 
\hat{\beta}_{k,i}.$ The main effect estimation generated by A2D2E, denoted by $\hat{f}_d^{\mathrm{A2D2E}}$, is then recovered by 
accumulating the increments across bins $\hat{f}_d^{\mathrm{A2D2E}}(t) = 
\sum_{k=1}^{J(t)} \hat{\Delta}_d^k + c,$ where $c$ is a centering constant chosen so that $\hat{f}_d^{\mathrm{A2D2E}}$ has mean zero over 
the evaluated locations. The full procedure is 
summarized in Algorithm~\ref{alg:A2D2E}.

\begin{algorithm}[H]\small
\caption{A2D2E}
\label{alg:A2D2E}
\begin{algorithmic}[1]
\Require Supervised prediction model $\hat{f}$, training 
data $\mathcal{D}_n$, number of bins $K$, cell width 
$\delta$, target variable index $d$

\State Let $\{z_d^k\}_{k=1}^{K+1}$ be bin boundaries 
on axis $d$
\State Initialize $\hat{f}_d^{\mathrm{A2D2E}}(z_d^1) 
\gets 0$
\For{$k = 1$ to $K$}
  \State Initialize $\hat{\beta}_{k} \gets 0$
  \For{$i \in I_d^k$} \Comment{\textbf{Step 1: 
  Estimate $\hat{\beta}_{k}$ via D-optimal design}}
    \State Construct local design 
    $\mathcal{S}_{\mathrm{A2D2E},i}$ via 
    \eqref{eq:si_selection}
    \State Compute $\hat{\beta}_{k,i}$ via 
    \eqref{eq:estimate_beta} and set 
    $\hat{\beta}_{k} \gets \hat{\beta}_{k} + 
    \hat{\beta}_{k,i}$
  \EndFor
  \State $\hat{f}_d^{\mathrm{A2D2E}}(z_d^{k+1}) \gets 
  \hat{f}_d^{\mathrm{A2D2E}}(z_d^k) + 
  (z_d^{k+1} - z_d^k)\,\frac{\hat{\beta}_{k}}{|I_d^k|}$
  \Comment{\textbf{Step 2: Accumulate increment}}
\EndFor
\State $c \gets \frac{1}{K}\sum_{k=1}^{K} 
\hat{f}_d^{\mathrm{A2D2E}}(z_d^{k+1})$
\Comment{\textbf{Step 3: Centre}}
\State \Return $\hat{f}_d^{\mathrm{A2D2E}}(z_d^k) - c$ 
for all $k = 1, \dots, K+1$
\end{algorithmic}
\end{algorithm}

\section{Theoretical Results}\label{sec:thm}

We establish the theoretical properties of A2D2E in two parts. First, we show that the population target of A2D2E converges to the ALE main effect as the partition becomes finer, establishing that A2D2E and ALE share the same estimand 
in the population limit. Second, we extend this to the realistic setting where only a surrogate $\hat{f}$ is available. We show the consistency of A2D2E as both $n$ and $K$ grow, a theoretical guarantee not provided by existing methods.

In Sec.~\ref{sec:a2d2e}, we estimate $\beta_k$ using the empirical distribution of $X_{\backslash d}$ within bin $k$, which aligns with the spirit of ALE. Specifically, for each observation $i \in I_d^k$, we estimate the per-observation local slope $\beta_{k,i}$ via~\eqref{eq:estimate_beta}, 
and aggregate across the bin $
\beta_{k} := \mathbb{E}[\beta_{k,i} \mid X_d \in [z_d^k, z_d^{k+1}]],$ which is estimated in practice by its sample analogue
$\hat{\beta}_{k}
=
\frac{1}{|I_d^k|}
\sum_{i\in I_d^k} \hat{\beta}_{k,i}.$
Based on this, we define the population main effect target of A2D2E below.

\begin{definition}[Uncentred A2D2E main effect]
\label{def:a2d2e}
For $d \in [p]$, consider a partition
$\{[z_d^k, z_d^{k+1}]\}_{k=1}^K$ of the support of $X_d$.
The uncentred A2D2E main effect function is
\[
g_{d,\mathrm{A2D2E},K}(t)
\equiv
\sum_{k=1}^{J(t)}
(z_d^{k+1}-z_d^k)\,\beta_k,
\]
where $\beta_k$ is the bin-level slope defined in~\eqref{eq:aggregation}.
\end{definition}

The following proposition shows that as the partition becomes finer, the A2D2E population target converges 
to the ALE main effect.
\begin{theorem}[Integral representation]\label{thm:thm1}
Suppose $f \in C^2(\mathcal{X})$, with $\partial f / \partial x_d$ continuous on $\mathcal{X}$, and
$$
m_d(z) := \mathbb{E}\left[\frac{\partial f}{\partial x_d}(X_d, X_{\setminus d}) \,\bigg|\, X_d = z\right]
$$
continuous in $z$ with marginal density $p_d(z) > 0$ on $\mathcal{S}_d$.
Let $h_K = \max_k(z_d^{k+1} - z_d^k) \to 0$ and $\delta \leq h_K$.
Then as $K \to \infty$,
$$
g_{d,\text{A2D2E},K}(t) \to \int_{x_{\min,d}}^{t} 
\mathbb{E}\left[\frac{\partial f}{\partial x_d}(X_d, X_{\setminus d}) 
\,\bigg|\, X_d = z\right] dz = g_{d,\text{ALE}}(t),
$$
with approximation error $O(h_K + \delta^2)$, where the $O(\delta^2)$ term 
arises from the $D$-optimal hypercube approximation to the local slope, 
and the $O(h_K)$ term arises from the Riemann sum discretization of the integral.
\end{theorem}

The proof (Appendix~\ref{app:proof_prop}) exploits the fact that the $D$-optimal hypercube design cancels quadratic terms in the Taylor expansion by symmetry, giving a second-order accurate approximation to the local slope. In practice, $f$ is unknown and replaced by a surrogate 
$\hat{f}$ trained on finite data, introducing two sources 
of error: a statistical error from finite-sample slope 
estimation, and a surrogate error from approximating $f$ 
by $\hat{f}$. We establish consistency of A2D2E as both 
$n$ and $K$ grow under the following assumption.

\begin{assumption}\label{assump}
The surrogate model $\hat{f}_n$ satisfies $\sup_{\mathbf{x} \in \mathcal{X}} |\hat{f}_n(\mathbf{x})| 
< \infty$ and $\|\hat{f}_n - f\|_{L^\infty} \to 0$ almost surely as $n \to \infty$.
\end{assumption}

\begin{theorem}[Consistency of A2D2E]
\label{thm:unified}
Let $p_d(\cdot)$ denote the marginal density of $X_d$ 
with compact support $S_d$ and $p_d(z) > 0$ for all 
$z \in S_d$. Suppose $\{(\mathbf{x}_i, y_i)\}_{i=1}^n$ 
are i.i.d., $\mathcal{X}$ is compact, $\delta > 0$ 
with all design points lying within $\mathcal{X}$, 
and Assumption~\ref{assump} holds. 
Then for fixed $K$, as $n \to \infty$:
\[
\left|\hat{g}_{d,\mathrm{A2D2E},K,n}^{\hat{f}}(t) - 
g_{d,\mathrm{A2D2E},K}(t)\right| \to 0 \quad \text{a.s.}
\]
where $g_{d,\mathrm{A2D2E},K}(t)$ is the population 
A2D2E main effect under ${f}$ as defined in 
Definition~\ref{def:a2d2e}, and 
$\hat{g}_{d,\mathrm{A2D2E},K,n}^{\hat{f}}(t)$ is 
its finite-sample estimator based on $n$ observations, and the surrogate model is $\hat{f}$. 
\end{theorem}

The proof in Appendix~\ref{app:proof_unified} bounds the statistical 
error via the strong law of large numbers and the 
surrogate error via Assumption~\ref{assump}, 
establishing almost sure convergence for fixed $K$. We further strengthen this result by establishing joint consistency 
as both $n$ and $K$ grow simultaneously, showing that the finite-sample estimator converges to the ALE population target under mild rate conditions on the bin width $h_{K_n}$, the hypercube radius $\delta_n$, and the 
surrogate approximation error $\|\hat{f}_n - f\|_{L^\infty}$.

\begin{corollary}[Joint Consistency of A2D2E]\label{cor:joint}
Under the conditions of Theorem~\ref{thm:unified} and Assumption~\ref{assump}, 
suppose $K = K_n \to \infty$ and $h_{K_n} \to 0$, $\delta_n \leq h_{K_n}$, and $\frac{\|\hat{f}_n - f\|_{L^\infty}}{\delta_n} \to 0$ a.s.
Then
\begin{equation}
    \hat{g}^{\hat{f}_n}_{d,\text{A2D2E},K_n, n}(t) \to g_{d, \mathrm{ALE}}(t) \quad \text{a.s.}
\end{equation}
\end{corollary}

\section{Empirical Studies}\label{sec:exp}

We demonstrate the effectiveness of the proposed method through performance evaluation, computational complexity analysis, and sensitivity analysis. Without specification, we set $K=40$ (following the default choice in the ALE paper \cite{apley2020visualizing}) and $\delta$ equals to the bin width (i.e. $\delta=1/40=0.025$). Following \cite{apley2020visualizing}, bin boundaries are defined by the quantiles of $X_d$, so that each bin contains approximately equal numbers of observations. To evaluate accuracy, we adopt the Overall Root Mean Squared Error (ORMSE), defined as the average per-dimension RMSE over a grid of 100 equally spaced points across all $p$ dimensions, $\text{ORMSE} = \frac{1}{p}\sum_{d=1}^{p}\sqrt{\text{MSE}_d}$, averaged over 100 independent replications. The gradient of the prediction model is computed via the exact gradient, and finite differences are used when the predictor is not differentiable. To ensure fairness of comparison, all estimated main effects, as well as the ground truth, are centered to have mean zero by subtracting their average values over the evaluation points.

\subsection{Performance Comparison}
We evaluate performance on several additive benchmark 
functions $f(\mathbf{x}) = \sum_{d \in [p]} h_d(x_d)$, 
summarized in Tab.~\ref{tab:benchmark_functions}, under settings 
where the prediction model is sensitive to OOD inputs. 
Under additivity, the estimands of PD and ALE-based 
methods coincide regardless of feature dependence, 
making comparisons among them fair; the M plot targets 
a different estimand and is included only as a reference 
(see Appendix~\ref{app:detail}).

\paragraph{Case study on $\mathcal{K}$-NN Regression}

\begin{table}[htbp]
\centering
\small
\caption{Additive benchmark functions used in the experiments. The input space is $[0,1]^p$. Here $u_d = 2x_d - 1$ and $\sigma(z) = 1/(1+e^{-z})$.}
\label{tab:benchmark_functions}
\begin{tabular}{cl}
\toprule
Function ($p$) & Definition  \\
\midrule
$f_0$ ($p=3$)  & $x_1 + x_2^2 + 0\cdot x_3$  \\
$f_1$ ($p=2$)  & $\sin(u_1) + u_2^2$  \\
$f_2$ ($p=4$) & $\sin(10u_1) + \sin(u_2) + (u_3^3-u_3) + \sigma(10u_4)$  \\
$f_3$ ($p=4$) & $\mathbf{1}(u_1>0)u_1^2 + |u_2|^{1/2}\operatorname{sign}(u_2) + \sin(\pi u_3/2) + u_4\log(|u_4|+1)$  \\
$f_4$ ($p=4$) & $10\sin(u_1) + 0.1u_2^2 + 5e^{-u_3^2} + 0.05u_4$  \\
$f_5$ ($p=8$) & $\sum_{k=1}^8 \left[\frac{1}{k}\sin(k\pi u_k) + k\cos\!\left(\frac{\pi u_k}{2k}\right)\right]$  \\
\bottomrule
\end{tabular}
\end{table}

$\mathcal{K}$-NN \cite{james2013introduction} is a well-known statistical machine learning method that is valued for its fast training and interpretability. However, because its prediction is computed as an average over the $\mathcal{K}$ nearest training samples, it can be sensitive to OOD inputs. For each simulation function, the corresponding $\mathcal{K}$-NN hyperparameters were tuned separately for each environment using 10-fold cross-validation on a sample of size $500D$ sampled from $[0,1]^p$ uniformly. The candidate number of neighbors ($\mathcal{K}$) ranged from 1 to $\min(50, 10p)$, while the weighting scheme was chosen from \textit{uniform} and \textit{distance}.

For each simulation function of dimension $p$, we sample 
$100 \times p$ observations from $[0,1]^p$, and add mean-zero noise with variance equal to $30\%$ of the signal variance (i.e.\ $R^2\approx0.8$). The first variable is drawn as $x_1\sim\mathrm{Unif}(0,1)$, and the remaining variables are generated under three dependence levels: \textbf{Independent}, where all variables are sampled independently from $\mathrm{Unif}(0,1)$; \textbf{Low dependence}, where $x_d = x_1 + \varepsilon_d$ with $\varepsilon_d\sim\mathcal{N}(0,0.1^2)$; and \textbf{High dependence}, where the same construction is used but with $\varepsilon_d\sim\mathcal{N}(0,0.05^2)$.

\begin{table}[htbp]
\centering
\small
\caption{Mean ORMSE ($\pm$ 95\% CI) with noise fraction 0.3. Bold denotes the best method per row; underlining denotes the second best.}
\label{tab:ormse_noise_0p3}
\resizebox{\textwidth}{!}{%
\begin{tabular}{llccccc}
\toprule
Dependence & f & PD & M & ALE & DALE & A2D2E \\
\midrule
\multirow{6}{*}{Independent}
 & f0 & \textbf{0.0306 $\pm$ 0.0007} & 0.1299 $\pm$ 0.0017 & 0.1093 $\pm$ 0.0033 & 0.1953 $\pm$ 0.0002 & \underline{0.0886 $\pm$ 0.0010} \\
 & f1 & \textbf{0.0637 $\pm$ 0.0015} & 0.2021 $\pm$ 0.0034 & 0.1448 $\pm$ 0.0072 & 0.4040 $\pm$ 0.0020 & \underline{0.1336 $\pm$ 0.0031} \\
 & f2 & \textbf{0.2125 $\pm$ 0.0021} & \underline{0.3421 $\pm$ 0.0043} & 0.3990 $\pm$ 0.0075 & 0.4835 $\pm$ 0.0002 & 0.3704 $\pm$ 0.0025 \\
 & f3 & \textbf{0.0925 $\pm$ 0.0018} & 0.3056 $\pm$ 0.0039 & 0.2928 $\pm$ 0.0075 & 0.5024 $\pm$ 0.0003 & \underline{0.2638 $\pm$ 0.0019} \\
 & f4 & \textbf{0.3578 $\pm$ 0.0077} & 1.4528 $\pm$ 0.0161 & 1.1314 $\pm$ 0.0250 & 1.5716 $\pm$ 0.0011 & \underline{0.8381 $\pm$ 0.0076} \\
 & f5 & \textbf{0.2014 $\pm$ 0.0005} & \underline{0.2350 $\pm$ 0.0022} & 0.2499 $\pm$ 0.0016 & 0.2666 $\pm$ 0.0000 & 0.2397 $\pm$ 0.0004 \\
\midrule
\multirow{6}{*}{Low}
 & f0 & \underline{0.0932 $\pm$ 0.0015} & 0.3639 $\pm$ 0.0024 & 0.1123 $\pm$ 0.0051 & 0.2006 $\pm$ 0.0062 & \textbf{0.0834 $\pm$ 0.0025} \\
 & f1 & 0.2306 $\pm$ 0.0031 & 0.3693 $\pm$ 0.0025 & \underline{0.1728 $\pm$ 0.0087} & 0.4147 $\pm$ 0.0106 & \textbf{0.1504 $\pm$ 0.0049} \\
 & f2 & \textbf{0.2930 $\pm$ 0.0043} & 0.6823 $\pm$ 0.0060 & 0.3514 $\pm$ 0.0079 & 0.4833 $\pm$ 0.0007 & \underline{0.3113 $\pm$ 0.0051} \\
 & f3 & \textbf{0.1720 $\pm$ 0.0037} & 1.3514 $\pm$ 0.0058 & 0.2922 $\pm$ 0.0111 & 0.5025 $\pm$ 0.0015 & \underline{0.1998 $\pm$ 0.0045} \\
 & f4 & 1.7226 $\pm$ 0.0080 & 4.1034 $\pm$ 0.0155 & \underline{1.4658 $\pm$ 0.0382} & 1.5712 $\pm$ 0.0020 & \textbf{1.3270 $\pm$ 0.0132} \\
 & f5 & 0.2508 $\pm$ 0.0008 & 0.8672 $\pm$ 0.0043 & \underline{0.2487 $\pm$ 0.0021} & 0.2664 $\pm$ 0.0001 & \textbf{0.2266 $\pm$ 0.0007} \\
\midrule
\multirow{6}{*}{High}
 & f0 & \underline{0.1161 $\pm$ 0.0015} & 0.3789 $\pm$ 0.0023 & 0.1220 $\pm$ 0.0059 & 0.1985 $\pm$ 0.0046 & \textbf{0.0907 $\pm$ 0.0041} \\
 & f1 & 0.2768 $\pm$ 0.0024 & 0.3893 $\pm$ 0.0020 & \underline{0.2111 $\pm$ 0.0095} & 0.4132 $\pm$ 0.0111 & \textbf{0.1956 $\pm$ 0.0067} \\
 & f2 & 0.3516 $\pm$ 0.0043 & 0.7958 $\pm$ 0.0051 & \underline{0.3367 $\pm$ 0.0090} & 0.4823 $\pm$ 0.0008 & \textbf{0.2967 $\pm$ 0.0063} \\
 & f3 & \textbf{0.2038 $\pm$ 0.0039} & 1.4248 $\pm$ 0.0058 & 0.2821 $\pm$ 0.0114 & 0.4977 $\pm$ 0.0015 & \underline{0.2060 $\pm$ 0.0071} \\
 & f4 & 1.9022 $\pm$ 0.0071 & 4.3050 $\pm$ 0.0141 & 1.6043 $\pm$ 0.0397 & \underline{1.5721 $\pm$ 0.0025} & \textbf{1.5175 $\pm$ 0.0183} \\
 & f5 & 0.2538 $\pm$ 0.0008 & 1.0303 $\pm$ 0.0040 & \underline{0.2451 $\pm$ 0.0027} & 0.2660 $\pm$ 0.0001 & \textbf{0.2217 $\pm$ 0.0012} \\
\bottomrule
\end{tabular}%
}
\end{table}

Tab.~\ref{tab:ormse_noise_0p3} reveals two key insights. First, PD achieves the lowest ORMSE under the independent setting, as expected, since no extrapolation bias is present; yet among all ALE-based methods, A2D2E outperforms ALE-based methods across the majority of settings, with the largest gains under high feature correlation. Second, the advantage of A2D2E grows with input dimension $p$, consistent with our discussion in Sec.~\ref{sec2}.

\begin{figure}[htbp]
    \centering
    \begin{subfigure}[t]{0.3\textwidth}
        \includegraphics[width=\linewidth]{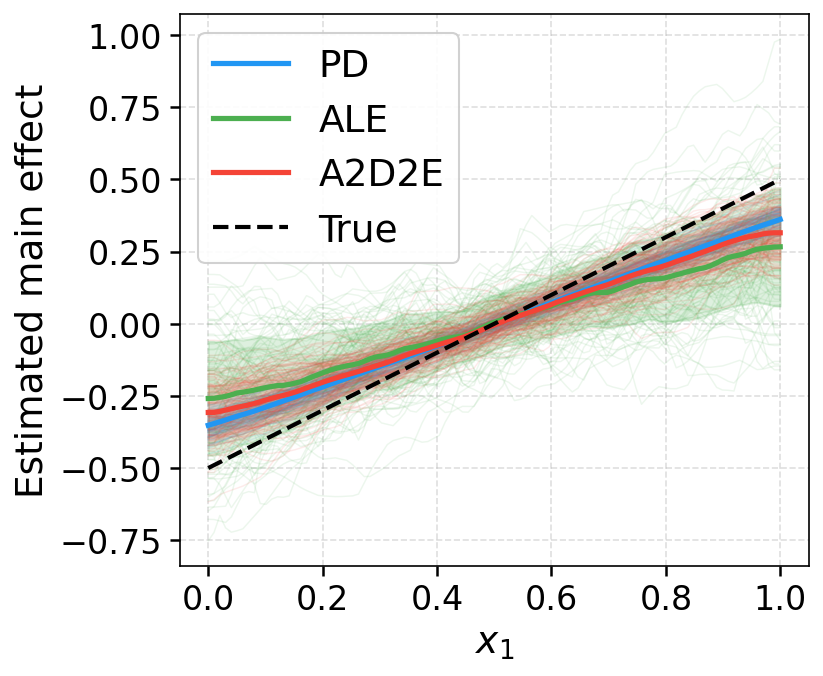}
        \subcaption{$x_1$}
    \end{subfigure}
    \hfill
    \begin{subfigure}[t]{0.3\textwidth}
        \includegraphics[width=\linewidth]{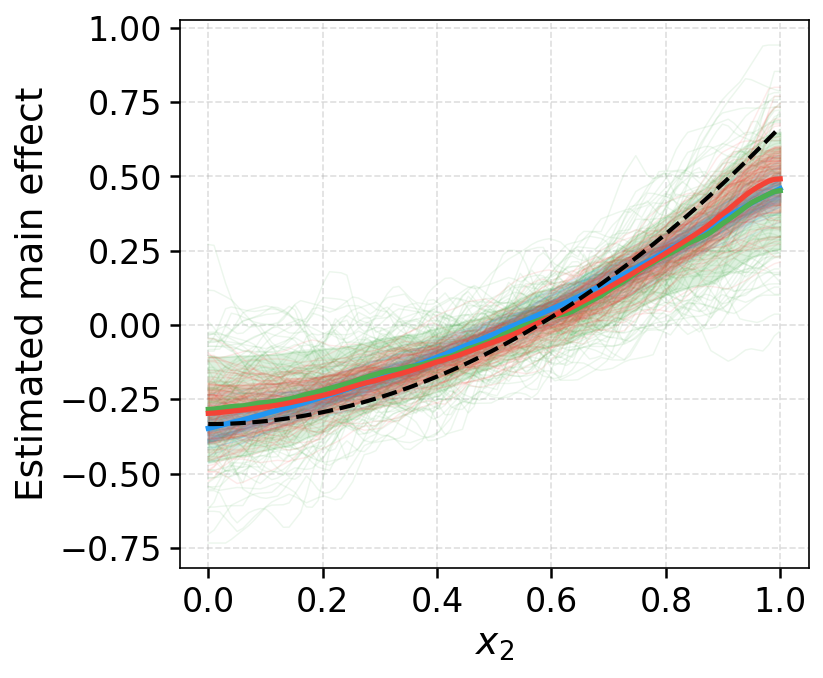}
        \subcaption{$x_2$}
    \end{subfigure}
    \hfill
    \begin{subfigure}[t]{0.3\textwidth}
        \includegraphics[width=\linewidth]{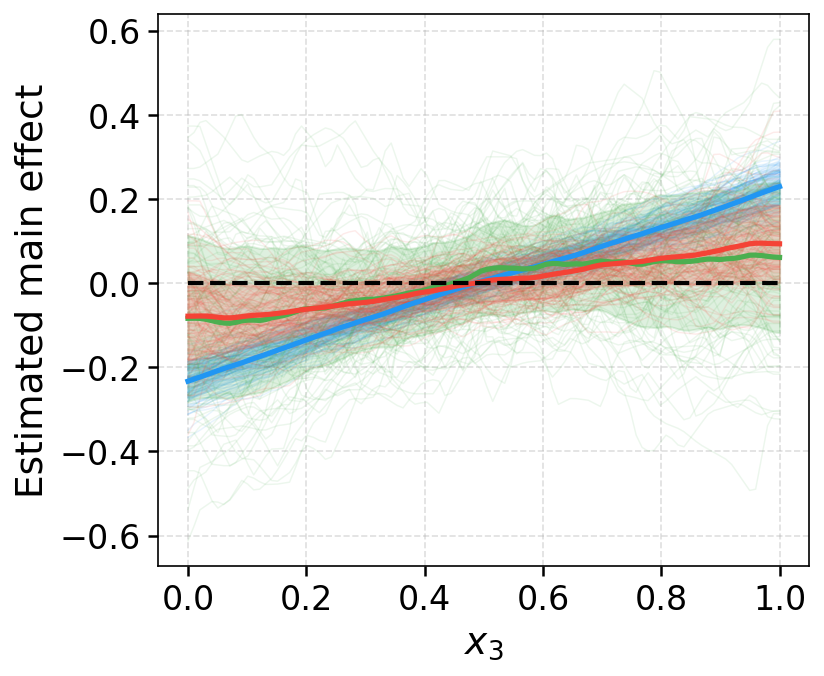}
        \subcaption{$x_3$ (null)}
    \end{subfigure}

    \caption{Estimated main effects for $f_0 = x_1 + x_2^2+0*x_3$ under low
    dependence with NN regression. Each panel shows all three methods:
    PD (\textcolor{pdblue}{\rule[0.5ex]{1em}{1.5pt}}), ALE (\textcolor{alegreen}{\rule[0.5ex]{1em}{1.5pt}}), and A2D2E (\textcolor{a2d2ered}{\rule[0.5ex]{1em}{1.5pt}}).
    Solid lines show the mean estimated effect across replications,
    shaded bands show the 95\% CI, faint lines show
    individual replications, and the \textbf{black dashed} line is
    the true effect.}
    \label{fig:f0_low_dep_main_effects}
\end{figure}

Fig.~\ref{fig:f0_low_dep_main_effects} illustrates the estimated main effects of PD, ALE and A2D2E for $f_0$ under low dependence (corresponding to the results in the 7-th row in Tab.~\ref{tab:ormse_noise_0p3}). All three methods recover the active effects of $x_1$ and $x_2$ reasonably well, though ALE exhibits the largest variability across replications. The null variable $x_3$ is most revealing, where PD shows a strong spurious linear trend due to extrapolation bias from correlated features. ALE reduces this but retains wide replication bands, while A2D2E remains closest to the true zero effect with the tightest uncertainty, confirming its robustness under feature correlation.

\begin{wrapfigure}{r}{0.5\textwidth}
    \centering
            \vspace{-1em}
    \includegraphics[width=\linewidth]{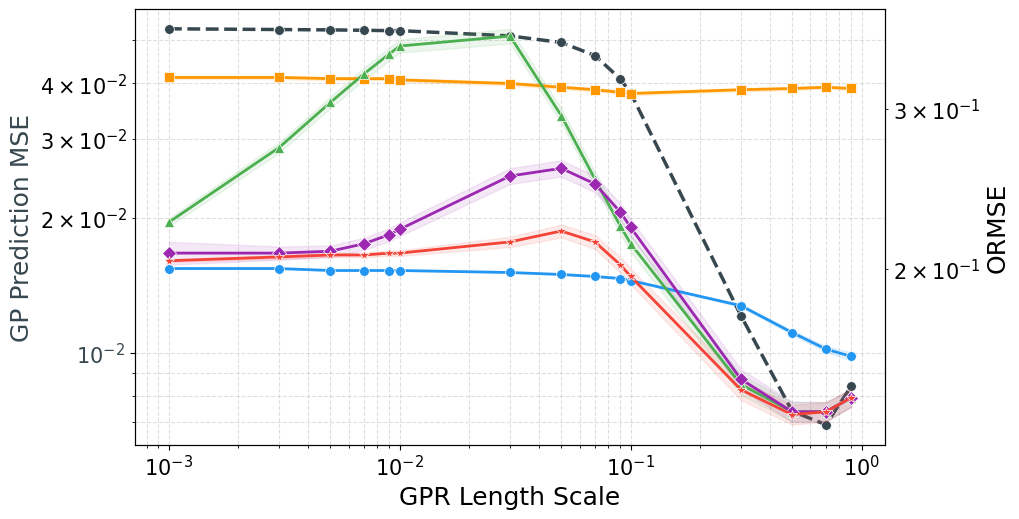}
    \caption{Case study on GPR with an RBF kernel across 20 length
    scales. The \textbf{black dashed} line shows the GP prediction MSE
    (left axis), while the right axis shows the ORMSE for
    PD (\textcolor{pdblue}{\rule[0.5ex]{1em}{1.5pt}}),
    M (\textcolor{morange}{\rule[0.5ex]{1em}{1.5pt}}),
    ALE (\textcolor{alegreen}{\rule[0.5ex]{1em}{1.5pt}}),
    DALE (\textcolor{daleviolet}{\rule[0.5ex]{1em}{1.5pt}}), and
    A2D2E (\textcolor{a2d2ered}{\rule[0.5ex]{1em}{1.5pt}}).
    Shaded bands show 95\% confidence intervals over 500 replications.}
    \label{fig:gp_case_study}
        \vspace{-1em}

\end{wrapfigure}

\paragraph{Case study on Gaussian Process (GP) Regression}
GP is a non-parametric predictor whose behaviour is governed by a kernel function; in the common RBF kernel, the length scale $\ell$ controls the smoothness and effective range of the fitted surface. A small $\ell$ produces a highly local, wiggly fit that interpolates the training data closely but generalizes poorly to unseen regions, while a large $\ell$ yields an oversmoothed surface that may fail to capture local structure. This sensitivity has a direct consequence for effect estimation methods that rely on evaluating the surrogate at OOD inputs. To investigate how GP length scale affects effect estimation on a locally sharp function, we consider $f(x_1, x_2, x_3) = \exp(-100(x_1 - 0.5)^2) + x_2+ 0*x_3$. We fit a GPR with an RBF kernel at 20 $l$, spanning from highly local to oversmoothed. The data is sampled using high dependence, each experiment is performed 500 times, and all other settings are the same as before.

The results reveal two notable strengths of A2D2E. First, compared to other ALE-based methods, A2D2E is the most stable across the entire length scale range, while ALE deteriorates sharply at moderate $l$ ($[0.03,0.05]$) and DALE exhibits high variance throughout the intermediate regime, A2D2E maintains consistently low ORMSE regardless of GP misspecification. Second, at the optimal length scale region ($[0.1,0.3]$), where the GP prediction MSE drops sharply and the model is well-calibrated, A2D2E achieves competitive performance with PD in this specific setting. 

\subsection{Sensitivity Analysis of $K$ and $\delta$}\label{sec:sensitivity}
We first examine how the performance of A2D2E changes with different $K$. We follow the sensitivity analysis of \cite{gkolemis2023dale}. The black-box function is defined as
\[
f(\mathbf{x}) = x_1 x_2 + x_1 x_3 - a\bigl((x_1-x_2)^2 - \tau^2\bigr)
\cdot \mathbf{1}[x_1 - x_2 \geq \tau]
+ a\bigl((x_1-x_2)^2 - \tau^2\bigr)
\cdot \mathbf{1}[x_2 - x_1 \geq \tau]
\]
with $\tau = 1.2$ and $a = 7$. The function is mild inside the region $|x_1 - x_2| < \tau$ but diverges rapidly outside it, creating a severely adversarial OOD setting. The training data consists of $1000$ observations. Following \cite{gkolemis2023dale}, no surrogate model is fitted, instead, we assume that the true function $fr$ is known and can be evaluated directly. The first feature $x_1$ is clustered around five centres $\{1.5, 3.0, 5.0,6.3, 8.2\}$ with standard deviation $0.1$, together with a small number of boundary points, covering the range $[0, 10]$. The second feature is generated as $x_2 = x_1 + \varepsilon_2$ with $\varepsilon_2 \sim \mathcal{N}(0, 0.6^2)$, inducing strong correlation with $x_1$. The third feature is an independent noise variable, $x_3 \sim \mathcal{N}(0, 20^2)$. Details of ground truth calculation can be found in Appendix~\ref{app:detail}.

\begin{table}[htbp]
\centering\small
\caption{Mean ORMSE ($\pm$ 95\% CI) over 100 independent replications with different $K$s. Bold denotes the best method per column.}
\label{tab:case2_ormse}
\resizebox{\textwidth}{!}{%
\begin{tabular}{lcccccccc}
\toprule
& \multicolumn{8}{c}{Number of bins $K$} \\
\cmidrule(lr){2-9}
Method & 1 & 2 & 3 & 4 & 5 & 10 & 20 & 40 \\
\midrule
$\hat{f}_{\text{ALE}}$   & 79.97 $\pm$ 0.36 & 37.96 $\pm$ 0.32 & 21.06 $\pm$ 0.28 & 15.14 $\pm$ 0.31 & 10.03 $\pm$ 0.24 & 4.64 $\pm$ 0.23 & 4.92 $\pm$ 0.36 & 5.00 $\pm$ 0.34 \\
$\hat{f}_{\text{DALE}}$  & 3.15 $\pm$ 0.14 & 1.81 $\pm$ 0.17 & 1.86 $\pm$ 0.16 & 1.68 $\pm$ 0.17 & 1.62 $\pm$ 0.16 & 3.09 $\pm$ 0.22 & 4.76 $\pm$ 0.35 & 5.01 $\pm$ 0.34 \\
$\hat{f}_{\text{A2D2E}}$ & \textbf{3.14 $\pm$ 0.14} & \textbf{1.80 $\pm$ 0.17} & \textbf{1.85 $\pm$ 0.16} & \textbf{1.67 $\pm$ 0.17} & \textbf{1.60 $\pm$ 0.16} & \textbf{3.08 $\pm$ 0.22} & \textbf{4.75 $\pm$ 0.35} & \textbf{4.99 $\pm$ 0.34} \\
\bottomrule
\end{tabular}%
}

\end{table}

 \begin{wrapfigure}{r}{0.4\textwidth}
    \centering
    \includegraphics[width=\linewidth]{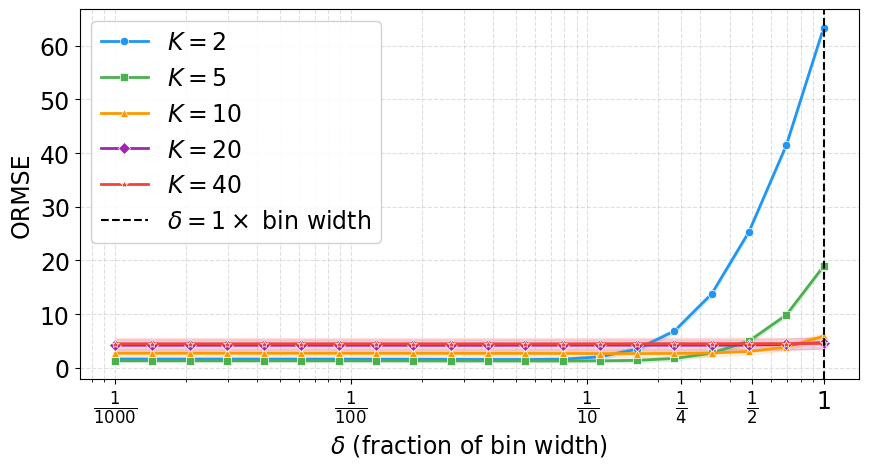}
    \caption{Sensitivity of A2D2E to the choice of $\delta$
    for varying $K \in \{2, 5, 10, 20, 40\}$, measured in
    ORMSE over 100 replications. The x-axis shows $\delta$
    as a fraction of the bin width. Performance is stable
    across a wide range of $\delta$ for moderate to large
    $K$ ($K \geq 10$), with deterioration only when $\delta$
    approaches the full bin width under very coarse binning
    ($K=2, 5$).}
    \label{fig:delta_sensitivity}
    \vspace{-2em}
\end{wrapfigure}

Tab.~\ref{tab:case2_ormse} shows that A2D2E consistently achieves the lowest ORMSE across all values of $K$, outperforming both ALE and DALE in every column. Notably, while ALE deteriorates sharply at small $K$ due to OOD sampling, A2D2E maintains stable and competitive performance across the full range of bin sizes, demonstrating its robustness to the choice of $K$.

To assess sensitivity to $\delta$, we fix $K$ and sweep $\delta$ from 
$\frac{1}{1000}$ to $1$ times the bin width over 100 replications. As shown in Fig.~\ref{fig:delta_sensitivity}, A2D2E remains stable across 
the full range for $K \geq 10$, with deterioration only at very coarse 
binning ($K = 2, 5$) when $\delta$ approaches the full bin width.

\paragraph{Additional Experiments and Limitations}
When the underlying function \(f\) is non-additive, the ground truth main effect conditioned on \(\mathcal{D}_n\) generally differs from that under uniformly sampled, uncorrelated inputs. As a result, benchmarking main effect estimation methods for a general non-additive \(f\) is intractable, and no reliable benchmark currently exists. In Appendix~\ref{app:sim}, we therefore report simulations on four non-additive functions, where the ground truth is defined under uncorrelated variables. Although this comparison is heuristic and not strictly fair, our method remains competitive under high correlation, with gains over ALE becoming more evident as the dependence among input variables increases.

To illustrate practical usefulness, we also include two real-data applications. In Appendix~\ref{app:real}, we study a regression task on the \texttt{Auto} dataset \citep{james2013introduction}, estimating the main effects of \emph{year}, \emph{acceleration}, \emph{horsepower}, and \emph{weight} on miles per gallon. We also consider a classification task on the Iris dataset \citep{fisher1936use}, visualizing the main effects of \emph{petal length}, \emph{petal width}, \emph{sepal length}, and \emph{sepal width} on the log-odds of class membership. We further provide a computational complexity comparison with existing baselines in Appendix~\ref{sec:comp}.

\section{Conclusion}
We proposed A2D2E, a model-agnostic estimator that leverages D-optimal 
design to minimize variance in main effect estimation, with theoretical 
guarantees of consistency. Extensive experiments 
confirm its advantage over ALE and DALE under feature correlation. 
The advantage of A2D2E is most pronounced when the surrogate 
model is sensitive to OOD evaluations, which can arise from two sources: the data sampling process, and the surrogate model itself. When neither source of OOD sensitivity is present, the gains of A2D2E over ALE are modest, and ALE remains a competitive and computationally cheaper alternative. A promising future direction is to develop a principled selection framework that guides practitioners in choosing among PD, ALE, and A2D2E based on observable data characteristics such as feature dependence strength and 
the out-of-distribution sensitivity of the surrogate model. While the D-optimal design provides a principled solution within the unified framework, it relies on local linearity and homoscedastic noise assumptions. Relaxing these, for instance through generalised optimal design criteria that accommodate nonlinear local models or heteroscedastic errors, represents a natural avenue for future work.
\section*{Acknowledgments}
This research was supported in part by Academia Sinica under Grant No.~AS-CDA-111-M05, and by the National Science and Technology Council (NSTC), Taiwan, under Grant Nos.~NSTC~113-2124-M-001-020 and NSTC~114-2628-M-001-008-MY3.

\bibliographystyle{plainnat}
\small
\bibliography{main}

@article{chernoff1953locally,
  title={Locally optimal designs for estimating parameters},
  author={Chernoff, Herman},
  journal={The Annals of Mathematical Statistics},
  pages={586--602},
  year={1953},
  publisher={JSTOR}
}

@article{gasser1984estimating,
  title={Estimating regression functions and their derivatives by the kernel method},
  author={Gasser, Theo and M{\"u}ller, Hans-Georg},
  journal={Scandinavian journal of statistics},
  pages={171--185},
  year={1984},
  publisher={JSTOR}
}

@article{muller1987bandwidth,
  title={Bandwidth choice and confidence intervals for derivatives of noisy data},
  author={M{\"u}ller, Hans-Georg and Stadtm{\"u}ller, Ulrich and Schmitt, Thomas},
  journal={Biometrika},
  volume={74},
  number={4},
  pages={743--749},
  year={1987},
  publisher={Oxford University Press}
}

@article{muller1984optimal,
  title={Optimal designs for nonparametric kernel regression},
  author={M{\"u}ller, Hans-Georg},
  journal={Statistics \& Probability Letters},
  volume={2},
  number={5},
  pages={285--290},
  year={1984},
  publisher={Elsevier}
}

@article{de2013derivative,
  title={Derivative estimation with local polynomial fitting},
  author={De Brabanter, Kris and De Brabanter, Joseph and Gijbels, Irene and De Moor, Bart},
  journal={Journal of Machine Learning Research},
  volume={14},
  number={1},
  pages={281--301},
  year={2013},
  publisher={MIT Press}
}

@book{james2013introduction,
  title={An introduction to statistical learning: with applications in R},
  author={James, Gareth and Witten, Daniela and Hastie, Trevor and Tibshirani, Robert},
  volume={103},
  year={2013},
  publisher={Springer}
}

@MISC{simulationlib, 
 author = {Surjanovic, S. and Bingham, D.}, 
 title = {Virtual Library of Simulation Experiments: Test Functions and Datasets}, 
 howpublished = {Retrieved September 18, 2025, from \url{http://www.sfu.ca/~ssurjano}} 
}

@book{wu2011experiments,
  title={Experiments: planning, analysis, and optimization},
  author={Wu, CF Jeff and Hamada, Michael S},
  year={2021},
  publisher={John Wiley \& Sons}
}

@article{kiefer1959optimum,
  title={Optimum experimental designs},
  author={Kiefer, Jack},
  journal={Journal of the Royal Statistical Society: Series B (Methodological)},
  volume={21},
  number={2},
  pages={272--304},
  year={1959},
  publisher={Wiley Online Library}
}

@article{brenning2023interpreting,
  title={Interpreting machine-learning models in transformed feature space with an application to remote-sensing classification},
  author={Brenning, Alexander},
  journal={Machine Learning},
  volume={112},
  number={9},
  pages={3455--3471},
  year={2023},
  publisher={Springer}
}

@article{hooker2007generalized,
  title={Generalized functional anova diagnostics for high-dimensional functions of dependent variables},
  author={Hooker, Giles},
  journal={Journal of computational and graphical statistics},
  volume={16},
  number={3},
  pages={709--732},
  year={2007},
  publisher={Taylor \& Francis}
}

@article{goldstein2015peeking,
  title={Peeking inside the black box: Visualizing statistical learning with plots of individual conditional expectation},
  author={Goldstein, Alex and Kapelner, Adam and Bleich, Justin and Pitkin, Emil},
  journal={journal of Computational and Graphical Statistics},
  volume={24},
  number={1},
  pages={44--65},
  year={2015},
  publisher={Taylor \& Francis}
}

@article{fathipour2023mean,
  title={Mean cutting force prediction of conical picks using ensemble learning paradigm},
  author={Fathipour-Azar, Hadi},
  journal={Rock Mechanics and Rock Engineering},
  volume={56},
  number={1},
  pages={221--236},
  year={2023},
  publisher={Springer}
}

@article{lundberg2017unified,
  title={A unified approach to interpreting model predictions},
  author={Lundberg, Scott M and Lee, Su-In},
  journal={Advances in neural information processing systems},
  volume={30},
  year={2017}
}

@article{hastie2017generalized,
  title={Generalized additive models},
  author={Hastie, Trevor J},
  journal={Statistical models in S},
  pages={249--307},
  year={2017},
  publisher={Routledge}
}

@article{moosbauer2021explaining,
  title={Explaining hyperparameter optimization via partial dependence plots},
  author={Moosbauer, Julia and Herbinger, Julia and Casalicchio, Giuseppe and Lindauer, Marius and Bischl, Bernd},
  journal={Advances in neural information processing systems},
  volume={34},
  pages={2280--2291},
  year={2021}
}

@article{tella2026advancing,
  title={Advancing Flood Susceptibility Mapping with Explainable AI: A Novel Application of Accumulated Local Effects (ALE)},
  author={Tella, Abdulwaheed and Pham, Quoc Bao and Zahidi, Izni and Fai, Chow Ming and Ibrahim, Karim Sherif Mostafa Hassan},
  journal={Water Resources Management},
  volume={40},
  number={4},
  pages={140},
  year={2026},
  publisher={Springer}
}

@article{gholami2024assessment,
  title={An assessment of global land susceptibility to wind erosion based on deep-active learning modelling and interpretation techniques},
  author={Gholami, Hamid and Mohammadifar, Aliakbar and Song, Yougui and Li, Yue and Rahmani, Paria and Kaskaoutis, Dimitris G and Panagos, Panos and Borrelli, Pasquale},
  journal={Scientific Reports},
  volume={14},
  number={1},
  pages={18951},
  year={2024},
  publisher={Nature Publishing Group UK London}
}

@article{charnigo2011generalized,
  title={A generalized C p criterion for derivative estimation},
  author={Charnigo, Richard and Hall, Benjamin and Srinivasan, Cidambi},
  journal={Technometrics},
  volume={53},
  number={3},
  pages={238--253},
  year={2011},
  publisher={Taylor \& Francis}
}

@article{welchowski2022techniques,
  title={Techniques to improve ecological interpretability of black-box machine learning models: case study on biological health of streams in the United States with Gradient Boosted Trees},
  author={Welchowski, Thomas and Maloney, Kelly O and Mitchell, Richard and Schmid, Matthias},
  journal={Journal of Agricultural, Biological and Environmental Statistics},
  volume={27},
  number={1},
  pages={175--197},
  year={2022},
  publisher={Springer}
}

@article{zhou2000derivative,
  title={On derivative estimation in spline regression},
  author={Zhou, Shanggang and Wolfe, Douglas A},
  journal={Statistica Sinica},
  pages={93--108},
  year={2000},
  publisher={JSTOR}
}

@article{apley2020visualizing,
  title={Visualizing the effects of predictor variables in black box supervised learning models},
  author={Apley, Daniel W and Zhu, Jingyu},
  journal={Journal of the Royal Statistical Society Series B: Statistical Methodology},
  volume={82},
  number={4},
  pages={1059--1086},
  year={2020},
  publisher={Oxford University Press}
}

@article{roy2025prediction,
  title={Prediction of mechanical properties of eco-friendly concrete using machine learning algorithms and partial dependence plot analysis},
  author={Roy, Tonmoy and Das, Pobithra and Jagirdar, Ravi and Shhabat, Mousa and Abdullah, Md Shahriar and Kashem, Abul and Rahman, Raiyan},
  journal={Smart Construction and Sustainable Cities},
  volume={3},
  number={1},
  pages={2},
  year={2025},
  publisher={Springer}
}

@article{ahnert2007numerical,
  title={Numerical differentiation of experimental data: local versus global methods},
  author={Ahnert, Karsten and Abel, Markus},
  journal={Computer Physics Communications},
  volume={177},
  number={10},
  pages={764--774},
  year={2007},
  publisher={Elsevier}
}

@article{shi2023clarifying,
  title={Clarifying relationship between PM2. 5 concentrations and spatiotemporal predictors using multi-way partial dependence plots},
  author={Shi, Haoze and Yang, Naisen and Yang, Xin and Tang, Hong},
  journal={Remote Sensing},
  volume={15},
  number={2},
  pages={358},
  year={2023},
  publisher={MDPI}
}

@article{knowles2014methods,
  title={Methods for numerical differentiation of noisy data},
  author={Knowles, Ian and Renka, Robert J},
  journal={Electron. J. Differ. Equ},
  volume={21},
  pages={235--246},
  year={2014}
}

@article{fisher1936use,
  title={The use of multiple measurements in taxonomic problems},
  author={Fisher, Ronald A},
  journal={Annals of eugenics},
  volume={7},
  number={2},
  pages={179--188},
  year={1936},
  publisher={Wiley Online Library}
}

@inproceedings{gkolemis2023dale,
  title={DALE: Differential Accumulated Local Effects for efficient and accurate global explanations},
  author={Gkolemis, Vasilis and Dalamagas, Theodore and Diou, Christos},
  booktitle={Asian Conference on Machine Learning},
  pages={375--390},
  year={2023},
  organization={PMLR}
}

@article{zhu2025unveiling,
  title={Unveiling the predictive power of machine learning in coal gross calorific value estimation: An interpretability perspective},
  author={Zhu, Wei and Xu, Na and Hower, James C},
  journal={Energy},
  volume={318},
  pages={134781},
  year={2025},
  publisher={Elsevier}
}

@article{hakkoum2024global,
  title={Global and local interpretability techniques of supervised machine learning black box models for numerical medical data},
  author={Hakkoum, Hajar and Idri, Ali and Abnane, Ibtissam},
  journal={Engineering Applications of Artificial Intelligence},
  volume={131},
  pages={107829},
  year={2024},
  publisher={Elsevier}
}
\clearpage
\appendix

\addtocontents{toc}{\protect\setcounter{tocdepth}{2}}
\tableofcontents

\section{Review of Slope Estimation Literature}\label{app:relatedwork}

While the primary goal of main effect estimation is to estimate how the response changes by varying individual variables, our unified formulation reveals that the core statistical task reduces to estimating a local slope (see Sec.~\ref{sec:design}). This connects our work to a rich literature in statistics on nonparametric slope and derivative estimation, which has been studied largely independently of the interpretability literature. For example, in local polynomial regression \citep{muller1987bandwidth, 
gasser1984estimating, zhou2000derivative, charnigo2011generalized}, the goal is to estimate the slope of an unknown regression function by fitting a local linear model in a neighborhood of each query point. In classical finite difference methods \citep{de2013derivative, knowles2014methods, ahnert2007numerical}, the goal is to approximate the derivative of a function directly from noisy observations, either with or without constructing an intermediate regression function. In optimal design for slope estimation \citep{muller1984optimal, chernoff1953locally}, the goal is to select evaluation locations that maximize the precision of the resulting slope estimator. While these works differ in their specific targets and assumptions, they share the common thread of recovering local slope information from function evaluations. Main effect estimation, however, presents a distinct set of constraints: the function is a fitted black-box model that is fixed and given; the observations are deterministic queries to $\hat{f}$; and the slope estimates must be aggregated and accumulated across bins to recover a globally coherent effect curve.

\section{Technical Results}
\label{app:theory}
\subsection{Proof of Theorem~\ref{thm:thm1}}
\label{app:proof_prop}

\begin{proof} Throughout this proof we work with the population version of the estimator, setting $\hat{f} = f$. The surrogate error introduced by replacing $f$ with $\hat{f}$ is handled separately in Theorem~\ref{thm:unified}.

For each bin \(k\), write $h_k := z_d^{k+1}-z_d^k$ and $h_K := \max_{1\le k\le K} h_k.$
Recall that $g_{d,\mathrm{A2D2E},K}(t)
=
\sum_{k=1}^{J(t)} h_k \beta_{k}.$ We show that $\beta_{k}
=
m_d(\xi_k)+O(h_k)+O(\delta^2)$
for some \(\xi_k\in I^k_d\). The result then follows by a Riemann sum argument.

From \eqref{eq:estimate_beta}, for a fix \(\mathbf x_i\), the A2D2E per-observation local slope estimator is
\[
\hat{\beta}_{k,i}=
\frac{1}{2^{p-2}\delta^2}
\sum_{s\in\{-1,+1\}^p}
\frac{\delta}{2}s_d\,
f\!\left(\mathbf x_i+\frac{\delta}{2}s\right).
\]
Since \(f\in C^2(\mathcal X)\), a second-order Taylor expansion of
\(f(  \mathbf x_i+\frac{\delta}{2}s)\) around \(  \mathbf x_i\) gives
\[
f\!\left(\mathbf x_i+\frac{\delta}{2}s\right)
=
f(\mathbf x_i)
+\frac{\delta}{2}s^\top \nabla f(\mathbf x_i)
+\frac{\delta^2}{8}s^\top H_f(\mathbf x_i)s
+O(\delta^3),
\]
uniformly over \(s\in\{-1,+1\}^p\), where \(H_f(\mathbf x_i)\) is the Hessian of \(f\) at \(\mathbf x_i\).

Substituting this expansion into $\hat{\beta}_{k,i}$, we analyze the terms by symmetry. First, the zeroth-order term vanishes because $\sum_{s\in\{-1,+1\}^p} s_d = 0.$ Second, for the first-order term,
\[
\sum_{s\in\{-1,+1\}^p} s_d s_j
=
\begin{cases}
0, & j\neq d,\\[4pt]
2^p, & j=d,
\end{cases}
\]
so
\[
\frac{1}{2^{p-2}\delta^2}
\sum_s
\frac{\delta}{2}s_d
\cdot
\frac{\delta}{2}s^\top \nabla f(\mathbf x_i)
=
\frac{1}{2^{p-2}\delta^2}\cdot \frac{\delta^2}{4}\cdot 2^p\, \frac{\partial f}{\partial x_d}(\mathbf x_i)
=
\frac{\partial f}{\partial x_d}(\mathbf x_i).
\]
Third, every term in $\sum_s s_d\, s^\top H_f(\mathbf x_i)s$ is a linear combination of monomials of the form \(s_d s_j s_\ell\), which contain an odd number of sign factors. By symmetry of the full hypercube, $\sum_{s\in\{-1,+1\}^p} s_d s_j s_\ell = 0,$ so the quadratic term vanishes. Finally, the Taylor remainder contributes $\frac{1}{2^{p-2}\delta^2}
\sum_s
\frac{\delta}{2}s_d \, O(\delta^3)
=
O(\delta^2).$ Therefore, we have $\hat{\beta}_{k,i}=f^d(\mathbf x_i)+O(\delta^2),$ uniformly in \(\mathbf x_i\in\mathcal{X}\).

In A2D2E, the population bin-level slope $\beta_k$ is the conditional 
expectation of the per-observation local slope over bin $k$. From previous justifications, we have $\hat{\beta}_{k,i} = \frac{\partial f}{\partial x_d}(\mathbf{x}_i) + O(\delta^2)$ 
uniformly in $\mathbf{x}_i$, so taking the conditional expectation over 
$X_d \in I_d^k$ gives
\[
\beta_k
=
\mathbb{E}\!\left[\hat{\beta}_{k,i}(X_d, X_{\backslash d})\mid X_d\in I_d^k\right]
=
\mathbb{E}\!\left[\frac{\partial f}{\partial x_d}(X_d, X_{\backslash d})
\mid X_d\in I_d^k\right]
+O(\delta^2),
\] where the expectation is taken over the joint distribution of 
$(X_d, X_{\backslash d})$ conditional on $X_d \in I_d^k$.

Now use the continuity of $m_d(\cdot)$. Since $I_d^k$ is an interval 
of width $h_k$, by the mean value theorem there exists some 
$\xi_k\in I_d^k$ such that
\[
\mathbb{E}\!\left[\frac{\partial f}{\partial x_d}(X_d,X_{\backslash d})
\mid X_d\in I_d^k\right]
=
m_d(\xi_k)+O(h_k).
\]
Therefore, $\beta_k = m_d(\xi_k)+O(h_k)+O(\delta^2).$ Substituting back into the definition of $g_{d,\mathrm{A2D2E},K}(t)$,
\[
g_{d,\mathrm{A2D2E},K}(t)
=
\sum_{k=1}^{J(t)} h_k \beta_k
=
\sum_{k=1}^{J(t)} h_k \bigl(m_d(\xi_k)+O(h_k)+O(\delta^2)\bigr)
=
\sum_{k=1}^{J(t)} h_k m_d(\xi_k)
+ O(h_K) + O(\delta^2),
\]
where we used $\sum_{k=1}^{J(t)} h_k \leq t - x_{\min,d}$ to bound 
the remainder terms. Since $h_K\to 0$ and $\delta\leq h_K\to 0$, 
both remainder terms vanish. Moreover,
\[
\sum_{k=1}^{J(t)} h_k m_d(\xi_k)
\longrightarrow
\int_{x_{\min,d}}^{t} m_d(z)\,\mathrm{d}z
\]
as a Riemann sum for the continuous function $m_d$. Therefore,
\[
g_{d,\mathrm{A2D2E},K}(t)
\longrightarrow
\int_{x_{\min,d}}^{t}
\mathbb{E}\!\left[\frac{\partial f}{\partial x_d}(X_d,X_{\backslash d})
\,\Big|\, X_d = z\right]\mathrm{d}z
=
g_{d,\mathrm{ALE}}(t),
\]
which completes the proof.
\end{proof}

\subsection{Proof of Theorem~\ref{thm:unified}}
\label{app:proof_unified}

\begin{proof}
We decompose the total error using the triangle inequality:
\begin{align}
  &\bigl|\hat{g}_{d,\mathrm{A2D2E},K,n}^{\hat{f}}(t)
   -g_{d,\mathrm{A2D2E},K}(t)\bigr| \nonumber\\
  &\;\leq\;
  \underbrace{
    \bigl|\hat{g}_{d,\mathrm{A2D2E},K,n}^{\hat{f}}(t)
    -g_{d,\mathrm{A2D2E},K}^{\hat{f}}(t)\bigr|
  }_{\text{Term 1: statistical error}}
  +
  \underbrace{
    \bigl|g_{d,\mathrm{A2D2E},K}^{\hat{f}}(t)
    -g_{d,\mathrm{A2D2E},K}(t)\bigr|
  }_{\text{Term 2: surrogate error}}.
  \label{eq:decomp_full}
\end{align}
Term~1 is the statistical error, which measures how well the 
finite-sample estimator $\hat{g}_{d,\mathrm{A2D2E},K,n}^{\hat{f}}(t)$ 
approximates its population counterpart $g_{d,\mathrm{A2D2E},K}^{\hat{f}}(t)$ 
for a fixed surrogate $\hat{f}$. This error vanishes as $n \to \infty$ 
by the law of large numbers, since for fixed $\hat{f}$ the per-observation 
slope estimates $\hat{\beta}_{k,i}$ are i.i.d.\ functions of $\mathbf{x}_i$ 
alone. Term~2 is the surrogate error, which measures the discrepancy 
between the population A2D2E main effect computed under the surrogate 
$\hat{f}$ and that computed under the true function $f$. This error 
vanishes as $n \to \infty$ because $\hat{f}$ converges uniformly to 
$f$ by Assumption~\ref{assump}. We bound each term in turn.
\paragraph{Term 1: Statistical error.}

Write the finite-sample estimator explicitly as
\[
  \hat{g}_{d,\mathrm{A2D2E},K,n}^{\hat{f}}(t)
  = \sum_{k=1}^{J(t)}h_k \cdot\hat{\beta}_{k},
\]
where $\hat{\beta}_k = \frac{1}{|I_d^k|}\sum_{i\in I_d^k}\hat{\beta}_{k,i}$
is the finite-sample bin-level slope using data from $\mathcal{D}_n$,
and the population target under $\hat{f}$ is
$g_{d,\mathrm{A2D2E},K}^{\hat{f}}(t)
  = \sum_{k=1}^{J(t)} h_k\beta_k.$

Throughout this argument, we treat $\hat{f}$ as a 
fixed deterministic function satisfying 
Assumption~\ref{assump}. Under this treatment, 
$\hat{\beta}_{k,i} = \hat{\beta}_{k,i}(\mathbf{x}_i, \hat{f})$ 
is a deterministic function of $\mathbf{x}_i$ alone. 
Since $\{(\mathbf{x}_i, y_i)\}_{i=1}^n$ are i.i.d., 
the quantities $\hat{\beta}_{k,i}$ are i.i.d.\ across 
$i$ for any fixed $\hat{f}$. The SLLN therefore applies 
directly. Because $\mathcal{X}$ is compact and
$\hat{f}$ is measurable, $\hat{\beta}_{k,i}$ is a bounded
function of $\mathbf{x}_i$: indeed,
\[
  \bigl|\hat{\beta}_{k,i}\bigr|
  \;\leq\;
  \frac{1}{2^{p-2}\delta^2}\cdot\frac{\delta}{2}
  \cdot 2^{p-1}\cdot\sup_{\mathbf{x}\in\mathcal{X}}|\hat{f}(\mathbf{x})|
  \;<\;\infty.
\]

Fix $K$ and $k$. Apply the SLLN to the
numerator and denominator of $\hat{\beta}_k$ separately. The terms
$\mathbf{1}(x_{i,d}\in(z_d^k,z_d^{k+1}])\cdot\hat{\beta}_{k,i}$
are i.i.d.\ (since $\mathbf{x}_i$ are i.i.d.) with finite expectation
$p_d((z_d^k,z_d^{k+1}])\cdot
\beta_k$.
By the SLLN:
\begin{align*}
  n^{-1}\sum_{i=1}^n
  \mathbf{1}\bigl(x_{i,d}\in(z_d^k,z_d^{k+1}]\bigr)
  \cdot\hat{\beta}_{k,i}
  &\xrightarrow{\mathrm{a.s.}}
  p_d\bigl((z_d^k,z_d^{k+1}]\bigr)
  \cdot\beta_k,\\[4pt]
  n^{-1}\sum_{i=1}^n
  \mathbf{1}\bigl(x_{i,d}\in(z_d^k,z_d^{k+1}]\bigr)
  &\xrightarrow{\mathrm{a.s.}}
  p_d\bigl((z_d^k,z_d^{k+1}]\bigr)
  \;>\;0,
\end{align*}
where positivity of the denominator limit follows from
$p_d(z)>0$ on $S_d$. By the continuous mapping theorem applied to
the ratio (with the denominator bounded away from zero almost surely
for all large $n$) $\hat{\beta}_k
  \xrightarrow{\mathrm{a.s.}}
  \beta_k.$ Since $J(t)$ is finite and fixed for fixed $K$, a finite sum of
almost-surely convergent sequences converges almost surely $ 
  \hat{g}_{d,\mathrm{A2D2E},K,n}^{\hat{f}}(t)
  \xrightarrow{\mathrm{a.s.}}
  g_{d,\mathrm{A2D2E},K}^{\hat{f}}(t),$ so $\text{Term~1}\to 0$ almost surely as $n\to\infty$.

\paragraph{Term 2: Surrogate error.}

Since $g_{d,\mathrm{A2D2E},K}^{\hat{f}}$ and
$g_{d,\mathrm{A2D2E},K}$ differ only through the choice of
function ($\hat{f}$ vs.\ $f$), we write
\[
  \bigl|g_{d,\mathrm{A2D2E},K}^{\hat{f}}(t)
  -g_{d,\mathrm{A2D2E},K}(t)\bigr|
  \leq
  \sum_{k=1}^{J(t)} h_k
  \cdot\bigl|\beta_k^{f}-\beta_k^{\hat{f}}|,
\]
where $\hat{\beta}_{k}^{\hat{f}}$ and $\hat{\beta}_{k}^{f}$ 
denote the per-bin slope from~\eqref{eq:estimate_beta} 
computed using $\hat{f}$ and $f$ respectively.
For each $\mathbf{x}$, using linearity of~\eqref{eq:estimate_beta}:
\[
  \hat{\beta}_{k,i}^{\hat{f}}(\mathbf{x})-\hat{\beta}_{k,i}^{f}(\mathbf{x})
  =
  \frac{1}{2^{p-2}\delta^2}
  \sum_{s\in\{-1,+1\}^p}
  \frac{\delta}{2}\,s_d\cdot
  \Bigl[\hat{f}\!\Bigl(\mathbf{x}+\tfrac{\delta}{2}s\Bigr)
       -f\!\Bigl(\mathbf{x}+\tfrac{\delta}{2}s\Bigr)\Bigr].
\]
Taking absolute values and bounding pointwise by
$\|\hat{f}-f\|_{L^\infty}$:
\begin{align*}
  \bigl|\hat{\beta}_{k,i}^{\hat{f}}(\mathbf{x})
  -\hat{\beta}_{k,i}^{f}(\mathbf{x})\bigr|
  &\leq
  \frac{1}{2^{p-2}\delta^2}\cdot\frac{\delta}{2}
  \cdot\|\hat{f}-f\|_{L^\infty}
  \cdot\!\!\sum_{s\in\{-1,+1\}^p}\!|s_d|
  =
  \frac{1}{\delta}\,\|\hat{f}-f\|_{L^\infty},
\end{align*}
where we used $\sum_{s\in\{-1,+1\}^p}|s_d|=2^{p-1}$.
The bound is uniform in $\mathbf{x}$, so taking conditional
expectations preserves it. Summing over bins and using
$\sum_{k=1}^{J(t)} h_k \leq t - x_{\min,d}$:
\[
  \text{Term~2}
  \leq
  \frac{t}{\delta}\,\|\hat{f}-f\|_{L^\infty}
  \;\to\; 0 \quad\text{as }n\to\infty\quad (\text{by Assumption~\ref{assump}}).
\]

For fixed $K$, combining~\eqref{eq:decomp_full} with the bounds
above gives
\[
  \bigl|\hat{g}_{d,\mathrm{A2D2E},K,n}^{\hat{f}}(t)
  -g_{d,\mathrm{A2D2E},K}(t)\bigr|
  \;\leq\;
  \text{Term~1} + \frac{t}{\delta}\,\|\hat{f}-f\|_{L^\infty}
  \;\xrightarrow{\mathrm{a.s.}}\; 0
\]
as $n\to\infty$, establishing the first claim.
 
\end{proof}

\subsection{Proof of Corollary~\ref{cor:joint}}

\begin{proof}
Apply the triangle inequality:
\begin{align}
\left|\hat{g}^{\hat{f}_n}_{d,\text{A2D2E},K_n,n}(t) - g_{d,\mathrm{ALE}}(t)\right| 
&\leq 
\underbrace{\left|\hat{g}^{\hat{f}_n}_{d,\text{A2D2E},K_n,n}(t) - g_{d,\text{A2D2E},K_n}(t)\right|}_{\text{Term 1: statistical error}} \\
&+ \underbrace{\left|g_{d,\text{A2D2E},K_n}(t) - g_{d,\mathrm{ALE}}(t)\right|}_{\text{Term 2: discretization error}}.
\end{align}
Term 1 vanishes almost surely as $n \to \infty$ for fixed $K_n$ by 
Theorem~\ref{thm:unified}, and the rate condition 
$\|\hat{f}_n - f\|_{L^\infty} / \delta_n \to 0$ ensures the surrogate 
error within Term 1 also vanishes. Term 2 vanishes as $K_n \to \infty$ 
by Theorem~\ref{thm:thm1}, since $h_{K_n} \to 0$ and 
$\delta_n \leq h_{K_n} \to 0$ imply the approximation error 
$O(h_{K_n} + \delta_n^2) \to 0$. Combining both terms completes the proof.
\end{proof}

\section{Computational Complexity}\label{sec:comp}

While all methods utilize the same scale of information --- namely, the supervised model 
and the training set --- the way they perform computations affects their efficiency. We 
analyze the time complexity of estimating the main effect function for a single variable 
$d$, and then discuss the amortized cost when estimating all $p$ dimensions simultaneously. 
Let $C$ denote the cost of querying the prediction model once.

The time complexity of a PD plot is $\mathcal{O}(KnC)$, 
since for each of the $K$ bins it queries the prediction model once per observation. For 
ALE, the complexity becomes $\mathcal{O}(2KnC)$, as it must query the prediction model 
at the two endpoints of the bin containing each observation. For A2D2E, although the 
method is formulated as a local regression problem, the D-optimal hypercube design yields 
a closed-form slope estimator that requires no matrix inversion. Specifically, as shown in 
(\ref{eq:estimate_beta}), the slope estimate at each observation reduces to a weighted sum 
of $2^p$ model evaluations,
\begin{equation*}
\hat{\beta}_{k,i} = \frac{1}{2^{p-2}\delta^2}
\sum_{s \in \{-1,+1\}^p} 
\frac{\delta}{2} [s]_d \cdot 
\hat{f}\!\left(\mathbf{x}_i + \frac{\delta}{2}s\right),
\end{equation*}
so the dominant cost per observation is simply $\mathcal{O}(2^p C)$. The overall 
complexity of A2D2E for a single dimension is therefore $\mathcal{O}(2^p Kn C)$, where 
the exponential factor $2^p$ reflects the number of hypercube vertices evaluated per 
observation.

A key computational advantage of A2D2E 
emerges when estimating main effects for all $p$ dimensions jointly. Because the $2^p$ 
hypercube vertex evaluations in (\ref{eq:estimate_beta}) do not depend on the target 
dimension $d$, they can be computed once and reused to extract the slope estimate for 
every dimension $d \in [p]$ at negligible additional cost. In contrast, PD requires 
$\mathcal{O}(KnC)$ model evaluations per dimension and ALE requires $\mathcal{O}(2KnC)$, 
so both scale linearly with $p$ when all dimensions are estimated. The total amortized 
cost of A2D2E across all $p$ dimensions is therefore $\mathcal{O}(2^p Kn C)$, independent 
of $p$ in the model evaluation term, offering a significant advantage over PD and ALE 
in settings where $p$ is moderate and $n$ is large.

\paragraph{Empirical Studies}
To empirically validate the theoretical complexity analysis, we benchmark the 
wall-clock time of PD, ALE, and A2D2E under two scenarios: (i) estimating the 
main effect of a \emph{single} variable $d=1$, and (ii) estimating main effects 
for \emph{all} $p$ variables simultaneously. The true function is the additive 
function $f(x_1, \ldots, x_p) = x_1 + x_2^2$, with the remaining dimensions 
having zero effect. The prediction model is a $1$-Nearest Neighbour ($1$-NN) 
regressor, chosen for its fast evaluation and sensitivity to out-of-distribution 
inputs. We generate $n = 500$ i.i.d.\ $\mathrm{Unif}[0,1]^p$ training samples, 
fix the number of bins to $K = 40$, and set $\delta$ equal to one bin width. 
We sweep over input dimensions $p \in \{2, 3, 4, 5, 6, 8\}$ and report the mean 
wall-clock time over 5 independent runs. For the all-dimensions scenario, A2D2E 
uses the vertex-caching strategy described above, where the $2^p$ hypercube 
vertex evaluations are computed once per observation and reused across all 
dimensions.

\begin{table}[ht]
\centering
\caption{Mean wall-clock time (seconds) for PD, ALE, and A2D2E under two 
scenarios. \textit{Left}: single variable ($d=1$) with fixed $p=4$ and varying 
$n$. \textit{Right}: all $p$ variables simultaneously with fixed $n=500$ and 
varying $p$, where A2D2E uses vertex caching. All experiments run on the HPC with single Intel Xeon Platinum 8358, 2.60\,GHz.}
\small
\begin{tabular}{|c|ccc|c|ccc|}
\hline
\multicolumn{4}{|c|}{\textbf{Vary $n$ (single dim, $p=4$)}} & 
\multicolumn{4}{c|}{\textbf{Vary $p$ (all dims, $n=500$)}} \\
\hline
$n$ & PD (s) & ALE (s) & A2D2E (s) & $p$ & PD (s) & ALE (s) & A2D2E (s) \\
\hline
100  & 0.012 & 0.006 & 0.048 & 2 & 0.122 & 0.056 & 0.071 \\
200  & 0.024 & 0.011 & 0.094 & 3 & 0.183 & 0.084 & 0.089 \\
500  & 0.061 & 0.028 & 0.241 & 4 & 0.244 & 0.112 & 0.132 \\
1000 & 0.122 & 0.055 & 0.483 & 5 & 0.305 & 0.140 & 0.213 \\
2000 & 0.245 & 0.111 & 0.965  & 8 & 0.488 & 0.224 & 0.645 \\
\hline
\end{tabular}

\label{tab:complexity}
\end{table}

Tab.~\ref{tab:complexity} confirms that all three methods scale linearly with 
$n$, consistent with their theoretical complexities. As expected, A2D2E is slower 
than PD and ALE for a single dimension by a factor of approximately $2^p/2 = 8$ 
relative to ALE, reflecting the ratio of model evaluations per observation 
($2^p$ vs $2$). The left part of Tab.~\ref{tab:complexity} shows that when estimating all 
dimensions simultaneously, PD and ALE scale linearly with $p$, while A2D2E 
remains competitive for small $p$ due to vertex caching, with the exponential 
factor only becoming the dominant cost at $p \geq 6$. In the typical use case 
where all main effects are required jointly and $p$ is small to moderate, 
A2D2E therefore offers a favourable trade-off between per-dimension overhead 
and amortised model evaluation cost.

\section{Additional Numerical Simulations}
\subsection{Simulation Details}\label{app:detail}

\paragraph{M plot estimand under additivity.}
Under the additivity assumption $f(\mathbf{x}) = \sum_{d \in [p]} h_d(x_d)$, the estimands of PD and ALE-based methods coincide regardless of the feature dependence structure. To see this, note that for an additive $f$, the partial derivative satisfies $\frac{\partial f}{\partial x_d} = h_d'(x_d)$, which depends only on $x_d$ and not on $x_{\backslash d}$. Therefore, the ALE main effect reduces to
\begin{align*}
f_d^{\text{ALE}}(x_d) = \int_{x_0}^{x_d} E\left[\frac{\partial f}{\partial x_d}(z, X_{\backslash d}) \,\Big|\, X_d = z\right] dz = \int_{x_0}^{x_d} h_d'(z)\, dz = h_d(x_d) - h_d(x_0),
\end{align*}
which is identical to the PD main effect $f_d^{\text{PD}}(x_d) = E[f(x_d, X_{\backslash d})] - c = h_d(x_d) + \sum_{j \neq d} E[h_j(X_j)] - c$, up to a centering constant. Hence, under additivity, the estimands of PD, ALE, and DALE all coincide, making comparisons among them fair regardless of the dependence structure among features.

The M plot, however, does not share this property. Its estimand is defined as 
\begin{align*}
f_d^{\text{M}}(x_d) = E[f(X_d, X_{\backslash d}) \mid X_d = x_d] = h_d(x_d) + \sum_{j \neq d} E[h_j(X_j) \mid X_d = x_d],
\end{align*}
which includes the conditional expectations $E[h_j(X_j) \mid X_d = x_d]$ for $j \neq d$. When features are correlated, these terms are non-trivial and introduce the combined effects of other variables into the estimated main effect of $x_d$. As a result, the M plot does not recover $h_d(x_d)$ even under additivity when features are correlated, and is therefore included only as a reference in our comparisons.

\paragraph{Ground truth calculation for the sensitivity analysis.}
For the sensitivity analysis of $K$ in Sec.~\ref{sec:sensitivity}, we follow the same experimental setup as \cite{gkolemis2023dale}. The black-box function is defined as
\begin{align*}
f(\mathbf{x}) = x_1 x_2 + x_1 x_3 - a\left((x_1 - x_2)^2 - \tau^2\right) \cdot \mathbf{1}[x_1 - x_2 \geq \tau] + a\left((x_1 - x_2)^2 - \tau^2\right) \cdot \mathbf{1}[x_2 - x_1 \geq \tau],
\end{align*}
with $\tau = 1.2$ and $a = 7$. The first feature $x_1$ is clustered around five centres $\{1.5, 3.0, 5.0, 6.3, 8.2\}$ with standard deviation $0.1$, together with a small number of boundary points covering the range $[0, 10]$. The second feature is generated as $x_2 = x_1 + \varepsilon_2$ with $\varepsilon_2 \sim \mathcal{N}(0, 0.6^2)$, inducing strong correlation with $x_1$. The third feature is an independent noise variable, $x_3 \sim \mathcal{N}(0, 20^2)$.

The ground truth main effect of $x_1$ is computed following the accuracy comparison procedure of \cite{gkolemis2023dale}. Specifically, since no surrogate model is fitted and the true function $f$ is assumed to be directly evaluable, the ground truth is obtained via numerical integration of the conditional expectation $E\left[\frac{\partial f}{\partial x_1}(x_1, X_{\backslash 1}) \mid X_1 = z\right]$ over the data-generating distribution. Within the mild region $|x_1 - x_2| < \tau$, the partial derivative simplifies to $\frac{\partial f}{\partial x_1} = x_2 + x_3$, so the local effect at $x_1 = z$ is $E[X_2 + X_3 \mid X_1 = z] = z$, since $E[X_2 \mid X_1 = z] = z$ and $E[X_3] = 0$. Outside the mild region, the quadratic penalty term contributes additional curvature, and the ground truth is computed numerically by Monte Carlo integration over the conditional distribution of $X_{\backslash 1}$ given $X_1 = z$, using a large number of samples drawn from the data-generating distribution described above.

\subsection{Additional Simulations on Non-additive Functions}\label{app:sim}
We acknowledge that evaluating the performance of effect function estimation remains rare, even in relatively simple settings such as estimating main effect functions. This may be due to the difficulty of extracting the true effect function from commonly used simulation functions. Moreover, the definition of the estimation target varies across different methods, making fair comparisons even more challenging.

In this work, we select commonly used simulation functions as benchmarks. The first two are from \cite{simulationlib}: \texttt{franke} ($D=2$),
\begin{align*}
f(x_1, x_2) &= \frac{3}{4}\exp\left(-\frac{(9x_1-2)^2}{4} - \frac{(9x_2-2)^2}{4}\right) \\
&+ \frac{3}{4}\exp\left(-\frac{(9x_1+1)^2}{49} - \frac{9x_2+1}{10}\right) \\
&+ \frac{1}{2}\exp\left(-\frac{(9x_1-7)^2}{4} - \frac{(9x_2-3)^2}{4}\right) \\
&- \frac{1}{5}\exp\left(-(9x_1-4)^2 - (9x_2-7)^2\right),
\end{align*}
and \texttt{branin} ($D=2$),
\begin{align*}
f(x_1, x_2) = a\left(x_2 - bx_1^2 + cx_1 - r\right)^2 + s(1-t)\cos(x_1) + s,
\end{align*}
where $a=1$, $b=\frac{5.1}{4\pi^2}$, $c=\frac{5}{\pi}$, $r=6$, $s=10$, and $t=\frac{1}{8\pi}$. The third benchmark is \texttt{simple} ($D=4$),
\begin{align*}
f(x_1, x_2, x_3, x_4) = x_1 x_2 - x_2 x_3 + x_4 x_1.
\end{align*}
These functions allow for analytical integration of the ground truth. We simulate $100D$ training data points for each experiment. To model measurement uncertainty, Gaussian noise with mean zero and variance set to 10\% of the response variance is added to the output for each function. Each experiment is repeated 50 times to quantify the variability and uncertainty associated with each method.

While we cannot extract the exact main effect function for the non-additive case when the observations are correlated, we benchmark all methods using the main effect function computed under the independent setting as the reference. We fully acknowledge that this comparison is inherently unfair, as the reference function does not account for the correlation structure present in the data. Nevertheless, this can be viewed as a measure of each method's ability to recover the main effect estimation given imperfect, correlated data.

For each function, we define the ground truth effect function as
\begin{equation}
f_{x_n}(x_n) = \int_{z_i = x_0}^{x_n} \int_{X_{\backslash i}} \frac{\partial f(z_i, z_{\backslash i})}{\partial z_i} \, dz_{\backslash i} \, dz_i,
\label{eq:truth}
\end{equation}
which serves as a reference for evaluating the proposed methods. We acknowledge that this ground truth may not correspond exactly to the target estimated by methods such as PD plot. Nevertheless, we include PD and ALE plots in our benchmark comparison to assess the strengths and limitations of each approach.

To assess the robustness of each method to correlations among variables, we consider three levels of dependence. For each setting, we generate \(100D\) samples. The first variable is drawn from a uniform distribution, \(x_1 \sim \mathrm{Unif}(0,1)\). The remaining variables are generated according to the specified dependence level, the same as the process described in the main paper:
\begin{enumerate}[label=(\roman*)]
    \item \textbf{Independent:} all remaining variables are sampled independently from \(\mathrm{Unif}(0,1)\);
    \item \textbf{Low dependence:} each variable is constructed as \(x_j = x_1 + \varepsilon_j\) with 
    \(\varepsilon_j \sim \mathcal{N}(0, 0.1^2)\);
    \item \textbf{High dependence:} same as (ii), but with \(\varepsilon_j \sim \mathcal{N}(0, 0.05^2)\).
\end{enumerate}

\begin{table*}[htbp]
\caption{Summary of average prediction ORMSE (mean $\pm$ 1.96 SE) using NN as the prediction model with confidence intervals. Bold for best results within a pair of functions and dependence level.}
\label{tab:simulationnn}
\centering
\small
\begin{tabular}{|l|c|c|c|c|}
\hline
\textbf{Functions} & \textbf{Dependence} & \textbf{PD} & \textbf{ALE} & \textbf{A2D2E} \\ 
\hline
\texttt{franke} & Independent & \textbf{0.078 $\pm$ 0.004} & 0.079 $\pm$ 0.005 & 0.079 $\pm$ 0.05 \\ 
 & Low dependent & 0.185 $\pm$ 0.04 & \textbf{0.113 $\pm$ 0.004} & 0.114 $\pm$ 0.004 \\ 
 & High dependent & 0.287 $\pm$ 0.05 & 0.173 $\pm$ 0.05 & \textbf{0.173 $\pm$ 0.005} \\ 
\hline
\texttt{branin} & Independent & \textbf{0.105 $\pm$ 0.008} & 0.137 $\pm$ 0.008 & 0.136 $\pm$ 0.008 \\ 
 & Low dependent & 0.471 $\pm$ 0.140 & 0.437 $\pm$ 0.024 & \textbf{0.434 $\pm$ 0.024} \\ 
 & High dependent & 2.100 $\pm$ 0.606 & 0.766 $\pm$ 0.088 & \textbf{0.732 $\pm$ 0.088} \\ 
\hline

\texttt{simple} & Independent & \textbf{0.013 $\pm$ 0.001} & 0.018 $\pm$ 0.001 & 0.015 $\pm$ 0.001 \\ 
 & Low dependent & 0.180 $\pm$ 0.046 & 0.065 $\pm$ 0.003 & \textbf{0.062 $\pm$ 0.002} \\ 
 & High dependent & 0.802 $\pm$ 0.185 & 0.175 $\pm$ 0.013 & \textbf{0.172 $\pm$ 0.012} \\ 
\hline
\end{tabular}
\end{table*}

The conclusions drawn from Tab.~\ref{tab:simulationnn} are consistent with those from the main paper. Specifically, when the data are independently sampled, PD performs best among the three methods, as its extrapolation limitations are less pronounced in this setting. As the dependence level among input variables increases, however, a clear advantage of A2D2E becomes evident. In this more challenging setting, A2D2E substantially improves over ALE and DALE in many correlated-feature settings, while remaining competitive with PD when extrapolation bias is limited.

\subsection{Additional Real Case Studies}\label{app:real}

\paragraph{Classification Task: The \texttt{iris} Dataset} To further showcase the effectiveness and generality of the proposed method in a diverse real-world context, we apply it to the classical \texttt{iris} dataset from \cite{fisher1936use}. We emphasize that the results presented in this section are purely illustrative in nature. Since the true main-effect functions in real-world data are unknown, any observed differences or similarities between methods cannot be used as grounds for systematic comparison. Rather, the sole purpose here is to demonstrate the practical utility of main-effect function estimation when applied to real-world classification problems. Any conclusions drawn from these visualizations should therefore be understood as heuristic observations, not rigorous evaluations.

\begin{figure}[ht]
    \centering
    \begin{subfigure}[t]{0.24\textwidth}
        \includegraphics[width=\linewidth]{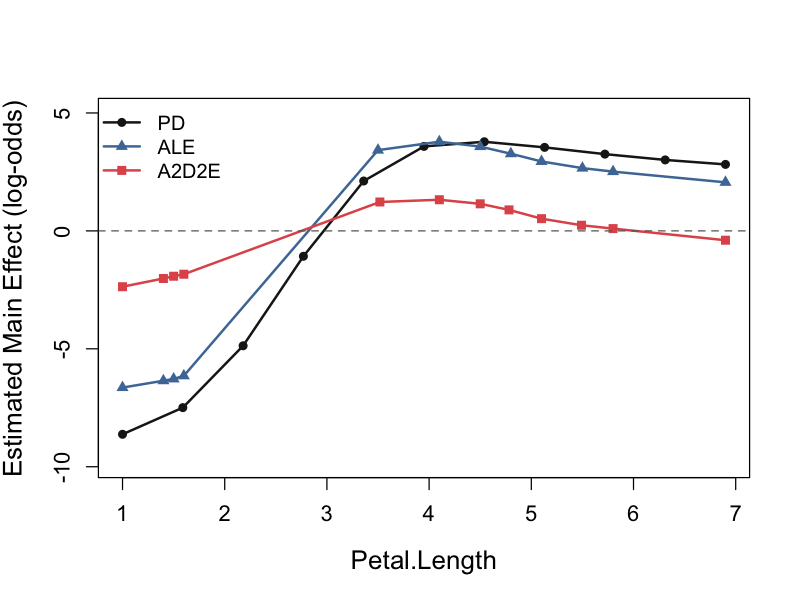}
    \end{subfigure}
    \hfill
    \begin{subfigure}[t]{0.24\textwidth}
        \includegraphics[width=\linewidth]{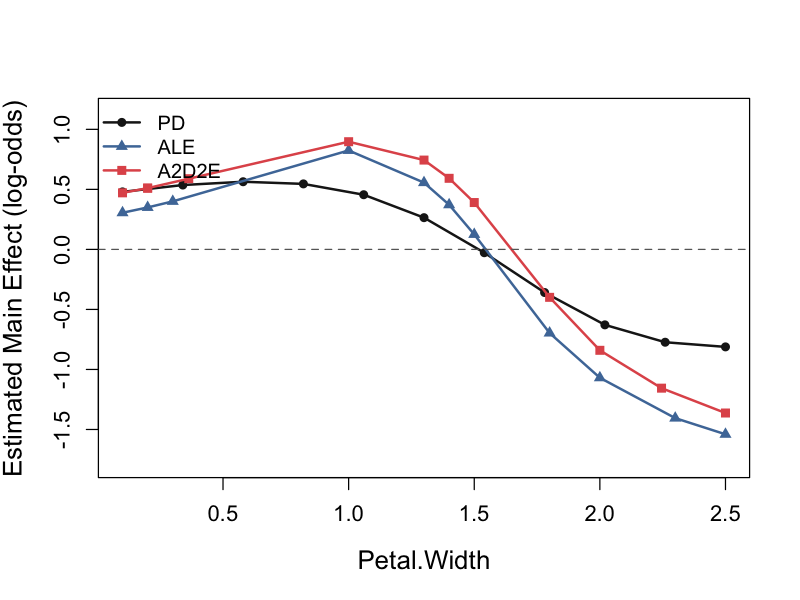}
    \end{subfigure}
    \hfill
    \begin{subfigure}[t]{0.24\textwidth}
        \includegraphics[width=\linewidth]{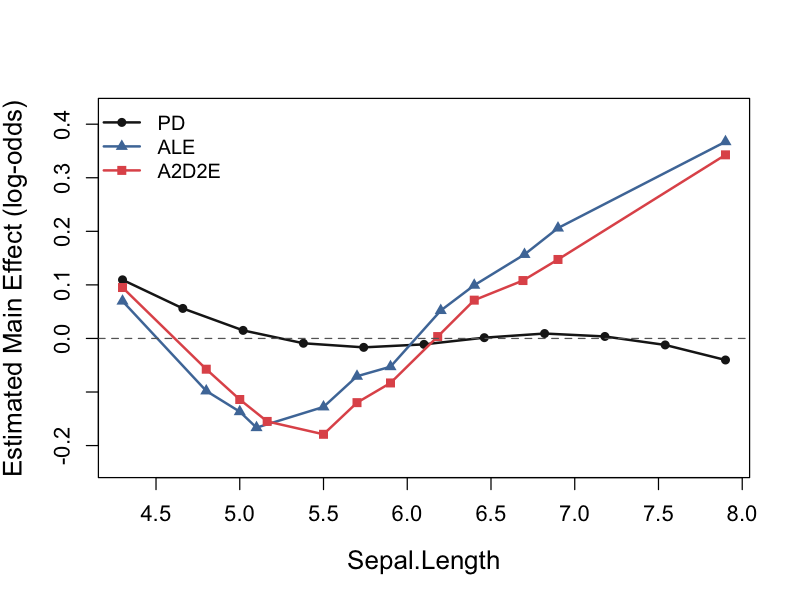}
    \end{subfigure}
    \hfill
    \begin{subfigure}[t]{0.24\textwidth}
        \includegraphics[width=\linewidth]{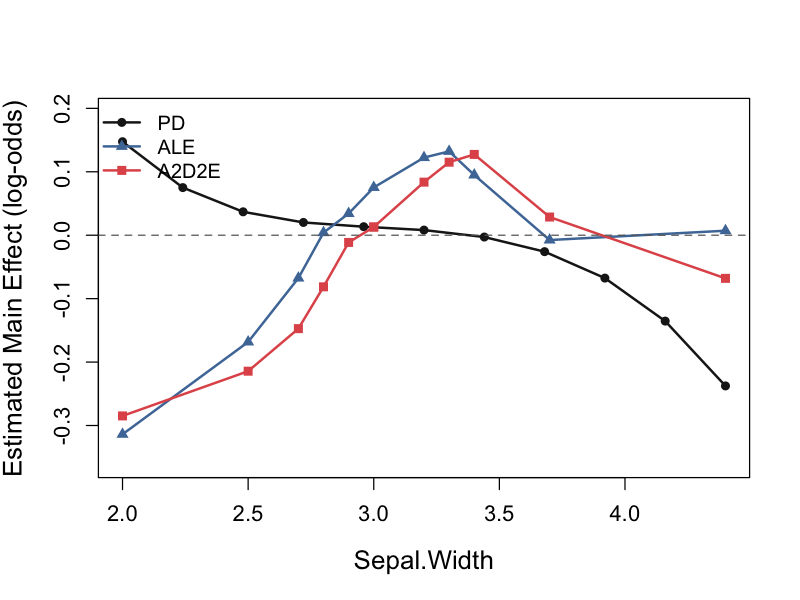}
    \end{subfigure}

    \caption{Estimated main-effects of the log-odds of classifying a sample as 
    \textit{versicolor} for the variables \emph{petal length}, \emph{petal width}, 
    \emph{sepal length}, and \emph{sepal width}, using PD, ALE, and the proposed 
    A2D2E algorithms.}
    \label{fig:iris}
\end{figure}
The \texttt{iris} dataset consists of three flower species (\textit{setosa}, \textit{versicolor}, and \textit{virginica}) characterized by four continuous features: sepal length, sepal width, petal length, and petal width. We visualize the estimated main-effect functions obtained using PD, ALE, and the proposed A2D2E methods in Fig.~\ref{fig:iris}.

To model the relationship between the input features and class probabilities, we trained a feedforward neural network with one hidden layer, implemented via the \texttt{nnet} package in \textsf{R}. A 10-fold cross-validation procedure was conducted to determine the optimal architecture and regularization strength. Specifically, the number of hidden units was searched over $\{4, 8, 12, 16\}$, and the $L_2$ weight-decay parameter was tuned over $\{0.0001, 0.001, 0.01\}$. The best configuration was found with eight hidden units and a decay of $0.01$, trained for up to 2000 iterations using the quasi-Newton optimization routine. This configuration achieved a cross-validated classification accuracy of approximately 97--98\%, indicating that the neural network successfully captured the nonlinear structure among the four features. To interpret the fitted model, we focused on visualizing the log-odds of predicting the \textit{versicolor} class relative to the reference class \textit{setosa}. For each feature, we estimated the corresponding main-effect function using PD, ALE, and A2D2E.

Fig.~\ref{fig:iris} illustrates how each feature influences the log-odds of predicting the class \textit{versicolor} relative to \textit{setosa} under the trained neural network. Among the four features, \textit{petal length} and \textit{petal width} exhibit the most dominant effects, showing sharp increases in the log-odds as their values increase from small to moderate levels, followed by a plateau where the classification confidence saturates. In contrast, \textit{sepal length} and \textit{sepal width} show weaker and more localized variations around their central ranges. We stress that these observations are heuristic in nature: without access to ground-truth main-effect functions, it is not possible to determine which method produces more accurate estimates, and the visual agreement or disagreement between methods should not be interpreted as evidence of superiority or inferiority.

These results nonetheless demonstrate the potential of the proposed A2D2E method as a practically useful, model-agnostic interpretability tool for complex classification tasks, and illustrate how main-effect estimation can be meaningfully extended beyond regression settings.

\begin{figure}[ht]
    \centering
    \begin{subfigure}[t]{0.24\textwidth}
        \includegraphics[width=\linewidth]{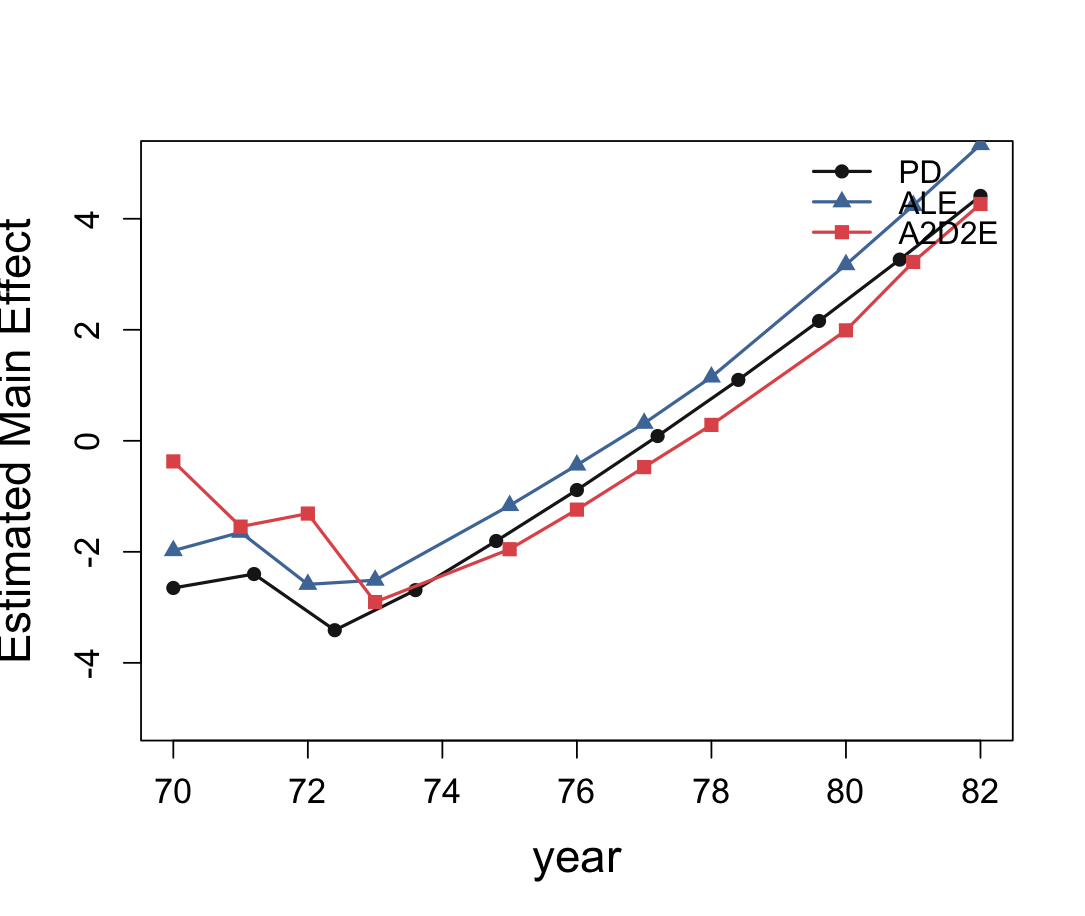}
    \end{subfigure}
    \hfill
    \begin{subfigure}[t]{0.24\textwidth}
        \includegraphics[width=\linewidth]{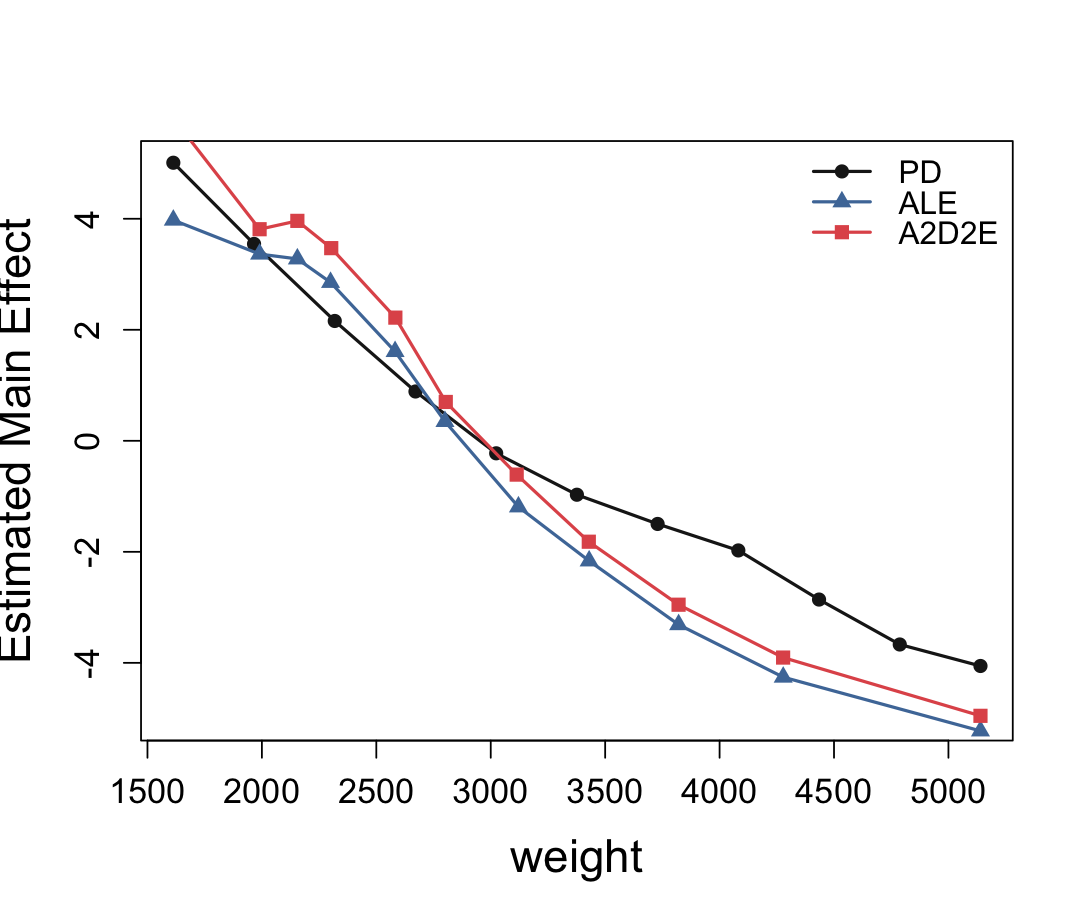}
    \end{subfigure}
    \hfill
    \begin{subfigure}[t]{0.24\textwidth}
        \includegraphics[width=\linewidth]{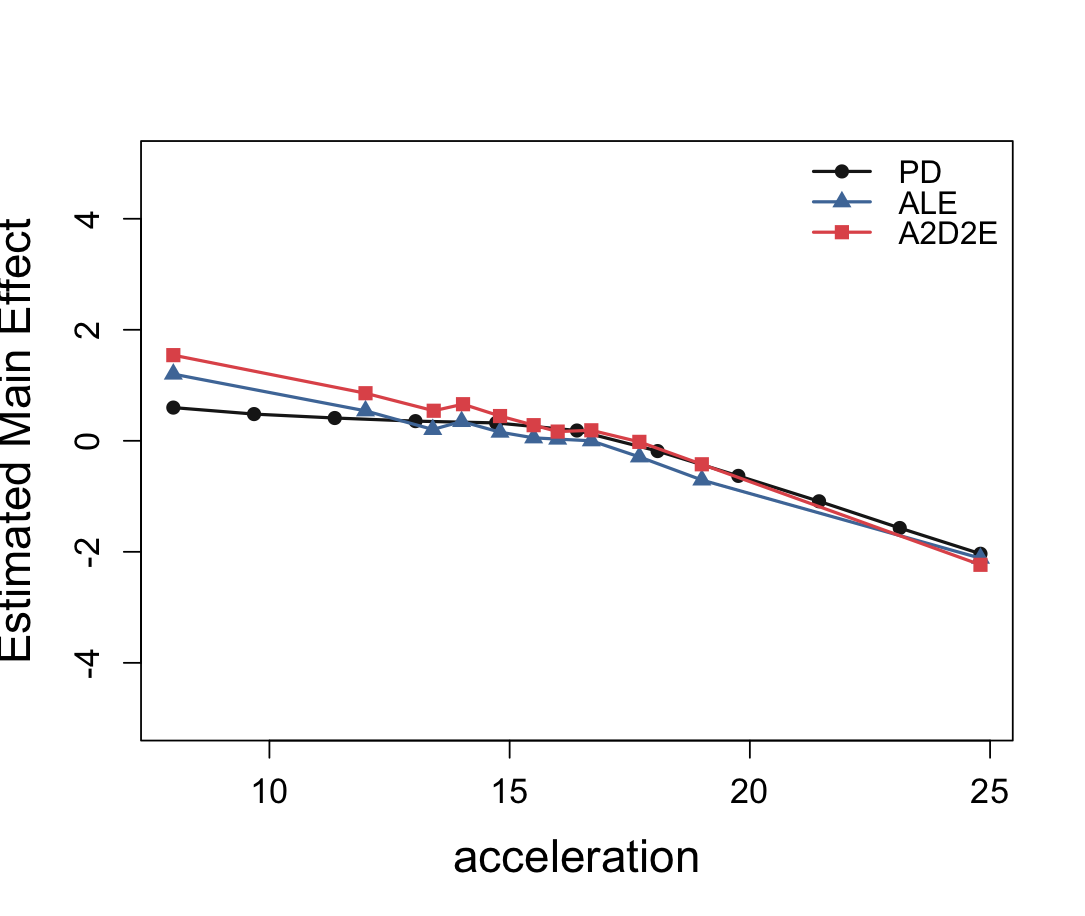}
    \end{subfigure}
    \hfill
    \begin{subfigure}[t]{0.24\textwidth}
        \includegraphics[width=\linewidth]{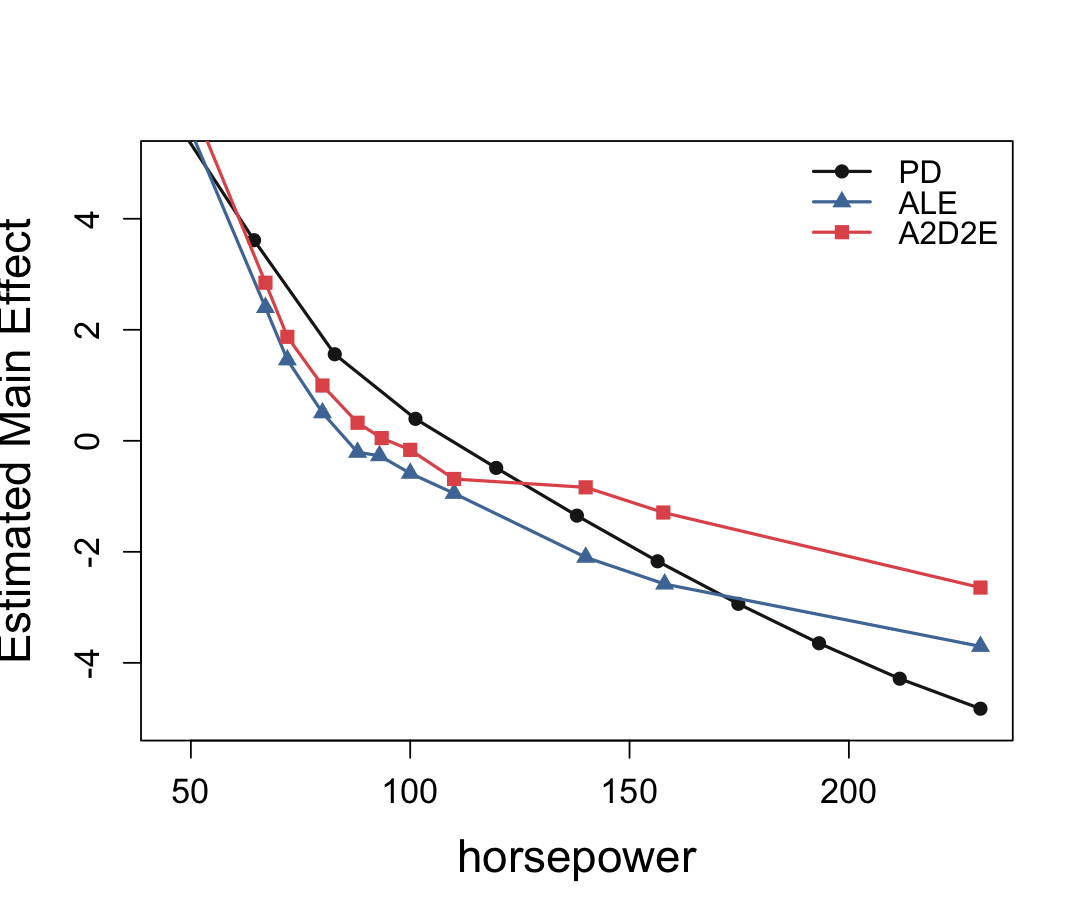}
    \end{subfigure}

    \caption{Estimated main-effect functions for the variables \emph{year}, 
    \emph{acceleration}, \emph{horsepower}, and \emph{weight} using PD, ALE, 
    and the proposed A2D2E algorithms.}
    \label{fig:auto}
\end{figure}

\paragraph{Regression Task: The \texttt{Auto} Dataset} We further apply the proposed method to a regression setting using the \texttt{Auto} dataset from \cite{james2013introduction}. As with the classification case study above, we stress that the absence of ground-truth main-effect functions in real-world data means that any observations made here are heuristic in nature and are not intended to serve as a basis for method comparison. The goal is solely to illustrate the practical utility of main-effect estimation in a real-world regression context.

We utilize a neural network with the same architecture settings as in the previous experiment to predict miles per gallon (MPG) using \emph{year}, \emph{acceleration}, \emph{horsepower}, and \emph{weight}. The visualization of the estimated main-effect functions for these variables is shown in Fig.~\ref{fig:auto}.

As shown in Fig.~\ref{fig:auto}, all methods exhibit broadly similar behavior across the four variables, with ALE and A2D2E producing more closely aligned shapes relative to PD. This may be attributed to the shared use of localization in both approaches. Interestingly, the estimated effect of \emph{year} suggests that vehicles tend to become more fuel-efficient over time, which is consistent with known trends in automotive technology. Examining \emph{horsepower} and \emph{acceleration}, ALE and A2D2E yield more similar results to each other than to PD. We believe this occurs because PD is known to perform less reliably when input variables are correlated; in this dataset, \emph{horsepower} and \emph{acceleration} exhibit a correlation of $-0.689$, which may explain the divergence of the PD estimates. We again caution that these observations are heuristic, and no definitive conclusions about method accuracy can be drawn in the absence of ground-truth main-effect functions.

\end{document}